\documentclass[lettersize,journal]{IEEEtran}
\usepackage{amsmath,amsfonts}
\usepackage{algorithmic}
\usepackage{algorithm}
\usepackage{array}
\usepackage[caption=false,font=normalsize,labelfont=sf,textfont=sf]{subfig}
\usepackage{textcomp}
\usepackage{stfloats}
\usepackage{url}
\usepackage{verbatim}
\usepackage{graphicx}
\usepackage{cite}
\begin{document}

\title{Controlling the Latent Diffusion Model for Generative Image Shadow Removal via Residual Generation}

\author{Xinjie Li, Yang Zhao, Dong Wang, Yuan Chen, Li Cao, Xiaoping Liu
\thanks{This work was supported by the National Natural Science Foundation of China under Grant 62277014.}
\thanks{Xiaoping Liu is the corresponding author of this work.}
\thanks{X. Li, Y. Zhao, D. Wang, L. Cao and X. Liu are with the School of Computers and Information, Hefei University of Technology, Hefei 230009, China (e-mail: 
xinjie\_li@mail.hfut.edu.cn, yzhao@hfut.edu.cn, Dongwang7@mail.hfut.edu.cn, lcao@hfut.edu.cn, liu@hfut.edu.cn}
\thanks{Y. Chen is with the School of Internet, Anhui University, Hefei 230039, China (e-mail: ychen@mail.ahu.edu.cn).}}

\markboth{Journal of \LaTeX\ Class Files,~Vol.~14, No.~8, August~2021}%
{Shell \MakeLowercase{\textit{et al.}}: A Sample Article Using IEEEtran.cls for IEEE Journals}



\maketitle

\begin{abstract}
Large-scale generative models have achieved remarkable advancements in various visual tasks, yet their application to shadow removal in images remains challenging. These models often generate diverse, realistic details without adequate focus on fidelity, failing to meet the crucial requirements of shadow removal, which necessitates precise preservation of image content. In contrast to prior approaches that aimed to regenerate shadow-free images from scratch, this paper utilizes diffusion models to generate and refine image residuals. This strategy fully uses the inherent detailed information within shadowed images, resulting in a more efficient and faithful reconstruction of shadow-free content. Additionally, to prevent the accumulation of errors during the generation process, a cross-timestep self-enhancement training strategy is proposed. This strategy leverages the network itself to augment the training data, not only increasing the volume of data but also enabling the network to dynamically correct its generation trajectory, ensuring a more accurate and robust output. In addition, to address the loss of original details in the process of image encoding and decoding of large generative models, a content-preserved encoder-decoder structure is designed with a control mechanism and multi-scale skip connections to achieve high-fidelity shadow-free image reconstruction. 
Experimental results demonstrate that the proposed method can reproduce high-quality results based on a large latent diffusion prior and faithfully preserve the original contents in shadow regions. 
\end{abstract}

\begin{IEEEkeywords}
Shadow removal, image generation, stable diffusion, image residual
\end{IEEEkeywords}

\section{Introduction}
\begin{figure}
    \centering
    \includegraphics[width=0.95\linewidth]{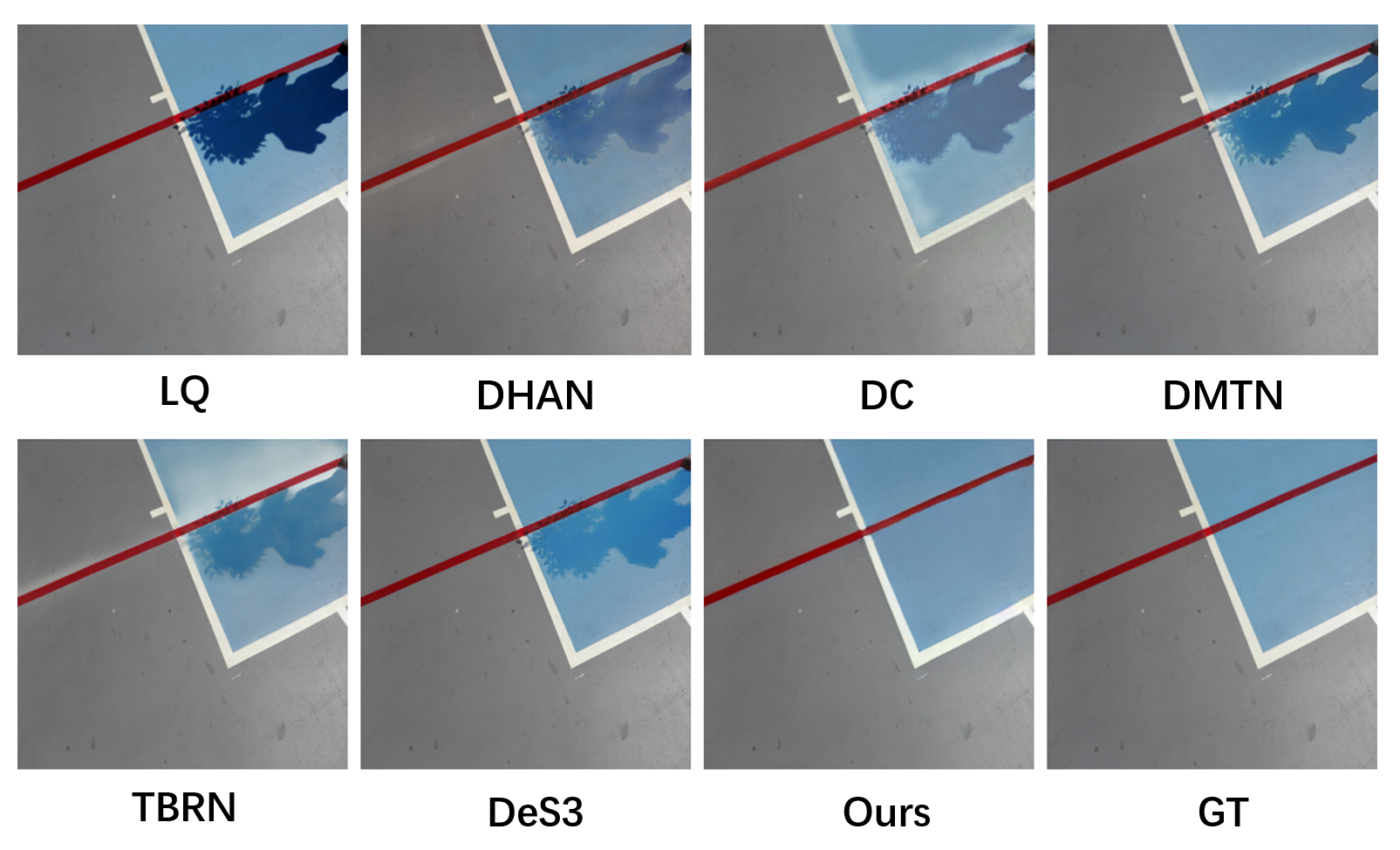}
    \caption{Current SOTA algorithms still cannot completely remove complex shadows. Owing to large-scale latent diffusion prior and the proposed residual generation diffusion, the proposed method can effectively remove shadows while faithfully preserving the image content.}
    \label{fig:ISTD+_example}
\end{figure}

\IEEEPARstart{S}{hadows} are an inherent part of our visual world, arising from the interplay of light and objects. While they contribute to the depth and realism of visual scenes, they may obscure important details, complicate object recognition, and lead to challenges in computer vision and image processing algorithms. Consequently, accurate shadow removal is of great importance in many fields, such as robotics, autonomous vehicles, medical imaging, and surveillance, which not only enhances the visual quality of images but also improves the performance of downstream applications by providing a clearer and more accurate representation of the scene.

With the development of deep neural networks (DNNs), DNN-based shadow removal algorithms have achieved significant progress \cite{guo2023shadowformer,li2023leveraging,liu2023shadow,xiao2024homoformer}. However, constrained by the network capacity and the absence of large-scale labeled shadow datasets, the state-of-the-art (SOTA) shadow removal algorithms still struggle to remove complex shadows completely, often leading to unnatural artifacts around the shadow boundaries. For example, Fig. \ref{fig:ISTD+_example} presents a challenging scene from the ISTD+ dataset \cite{le2019shadow}, in which many SOTA methods failed to effectively remove the human shadows, leading to unnatural artifacts. Recently, the large models, usually grounded in denoising diffusion probabilistic modeling, have demonstrated remarkable capabilities in photorealistic image generation task \cite{ho2020denoising,song2020denoising,rombach2022high,zhang2023adding,TMMstyle,TMManimegen} and low-level vision tasks such as super-resolution \cite{lin2023diffbir}, denoising \cite{xie2023diffusion}, and restoration \cite{kawar2022denoising,luo2023controlling,liu2024residual}. They excel at producing realistic texture details from noise, offering a generative paradigm for restoring visual features obscured by shadows.

Motivated by recent generative restoration and enhancement models \cite{lin2023diffbir,luo2023controlling}, this paper tends to leverage a pre-trained generative diffusion model for shadow removal. These pre-trained large-scale models, usually trained on massive datasets like Open Images \cite{kuznetsova2020open} and LAION \cite{schuhmann2022laion} for image generation, are capable of capturing high-level semantic features of images, facilitating an understanding of the image content and promoting shadow removal performance. However, directly applying these large diffusion models to remove shadows suffers from the generative hallucination problem. This phenomenon is quite common in generative image restoration models. For instance, DiffBIR \cite{lin2023diffbir} can restore visually sharper edges and clearer textures, but the generated textures may not be consistent with the ground truth. This is due to the fact that in order to achieve better generalization and diversity, the sampling and iteration processes within the diffusion models may lead to error accumulation, causing the generation process to deviate from the ideal trajectory gradually. However, fidelity is particularly crucial for the shadow removal task, as the regions outside the shadow should be maintained strictly, and the enhanced shadow areas should also be consistent with the original contents within the shadow. 

\begin{figure}
    \centering
    \includegraphics[width=0.95\linewidth]{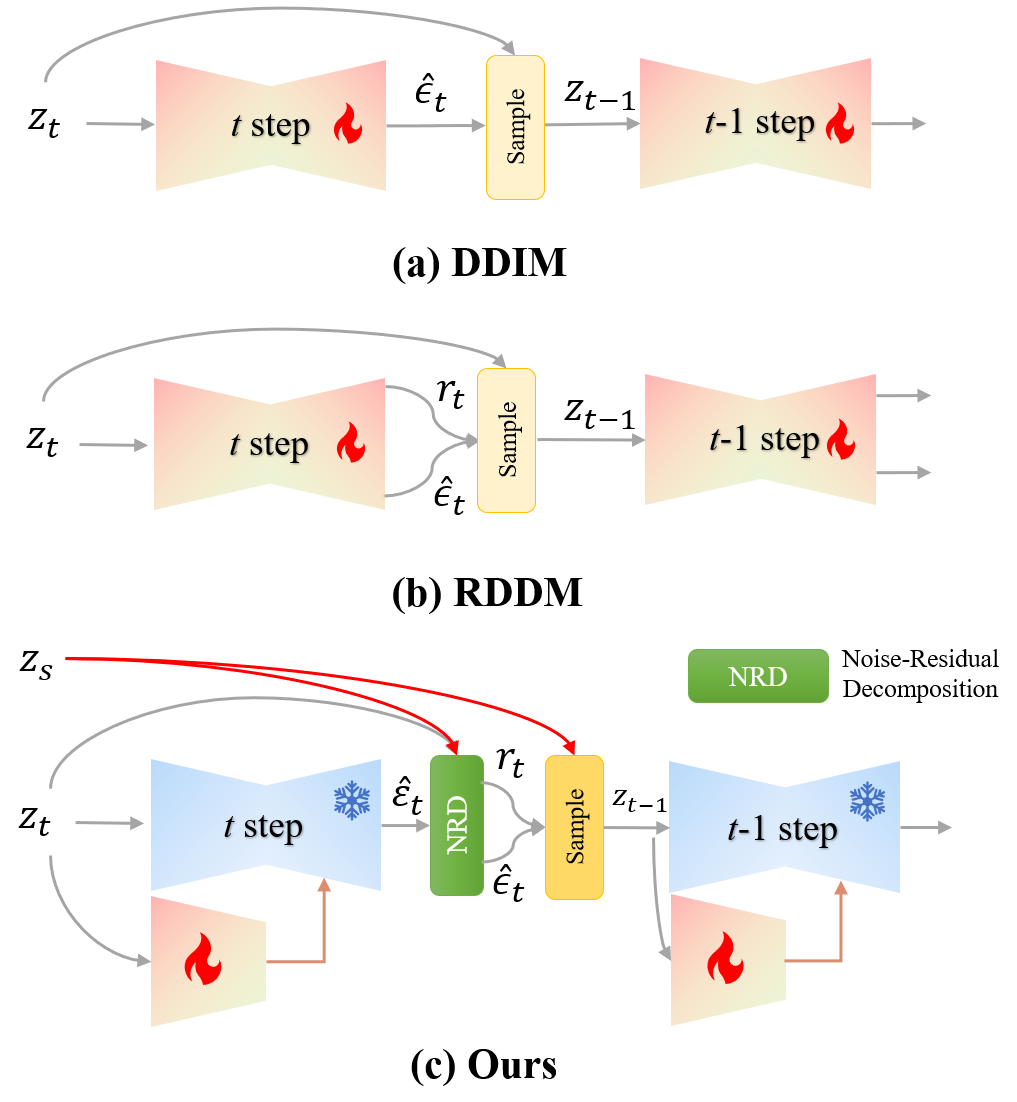}
    \caption{Diffusion backward processes of different methods. (a) Denoising Diffusion Implicit Models (DDIM). (b) Residual Denoising Diffusion Models (RDDM). (c) the proposed residual generation model.}
    \label{fig:DDIM-RDDM}
\end{figure}

To address the challenges above, this paper proposes a shadow removal model based on the residual generation and refinement process and latent diffusion prior. A residual generation diffusion model and corresponding training strategy are specifically designed for shadow removal tasks. In addition, the image encoder and decoder are improved to preserve the original contents faithfully. Compared to typical diffusion processes (DDIM \cite{song2020denoising} and RRDM \cite{liu2024residual}), as shown in Fig. \ref{fig:DDIM-RDDM}, the proposed approach makes minimal modifications to the pre-trained weights of the diffusion model and avoids training with randomly initialized branches to prevent the degradation of the model's capabilities. Furthermore, dense connections with the latent representations of shadow images are introduced during the generation and refinement of residuals, which further guide the diffusion generation process to retain image details.  

The contributions of this paper can be summarized as follows:
\begin{enumerate}
    \item  This paper introduces a new framework that fine-tunes a pre-trained large-scale generative model to generate and refine the shadow residual of the image, rather than re-generating the shadow-free image itself. This approach effectively mitigates the loss of fine details, enabling high-fidelity shadow removal.
    \item To address the issue of high-frequency information loss and alteration that occurs during the encoding and decoding processes used in large-scale generative models, a content-preserved encoder and decoder are proposed. Without compromising the original decoder's reconstruction capabilities, we introduce a controller for fine-tuning and training it to achieve high fidelity.
    \item To mitigate the accumulation of errors during the diffusion process, a cross-timestep self-enhancement strategy is proposed. By harnessing the network to generate its own training data, we achieve data augmentation while endowing each step of the network with the ability to correct the generative process.
    \item Extensive qualitative and quantitative experiments demonstrate that the proposed method can effectively leverage pre-trained large generative models to remove shadows from images and produce high-fidelity reconstruction outcomes. 
\end{enumerate}

\section{Related works}
\subsection{Shadow Removal}
Traditional shadow removal methods have typically relied on hand-crafted prior knowledge, such as assumptions about lighting conditions \cite{RN5,RN15}, gradient priors \cite{RN4,RN16}, and region-based characteristics \cite{RN6,RN17}. Though these methods can be effective in specific scenarios, their results are prone to artifacts and inconsistencies when the scene strays from the assumptions on which they were designed. This leads to a performance that is less than ideal in real-world applications where conditions may vary significantly from the expected idealized environments.

Recently, data-driven approaches have been developed to map shadowed images to shadow-free images automatically. Given the complexity and diversity of shadow scenarios, some methods \cite{wang2018stacked,fu2021auto,TMMshaodwremoval,xiao2024homoformer} rely on pre-obtained masks to locate shadows, thereby focusing on the removal of shadows in the masked areas. While these methods can brighten the designated areas, obtaining precise shadow masks poses another challenge, especially for soft shadows with unclear boundaries \cite{jin2024des3}. This limitation reduces the flexibility of these methods in adapting to different scenarios. Another category of methods does not rely on input masks but learns to identify shadow areas during the training process and takes additional measures to improve the accuracy of shadow estimation. For instance, Hu et al. \cite{hu2019mask} developed a method using unpaired data to avoid extensive annotations. Cun et al. \cite{cun2020towards} generated a large number of shadow-shadow-free image data pairs using a generative adversarial network to enhance training data. Jin et al. \cite{jin2021dc} proposed an unsupervised method that incorporates an image classification task as an auxiliary task to strengthen the network's attention to shadow areas. Liu et al. \cite{liu2023decoupled} further developed a multi-task estimation method to utilize the information contained in shadow data pairs fully. However, these methods still face the challenge of insufficient diversity in shadow scenarios, often leading to residual shadows or artifacts in the outputs.

\subsection{Diffusion-based Image Restoration}
Advancements in diffusion models have marked significant progress across visual tasks \cite{TMMinpainting,TMMrestore}. Researchers have increasingly turned to diffusion models to restore rich textural details in the shadowed regions of images. These methods primarily involve retraining diffusion models on the shadow datasets to improve image fidelity and generate shadow-free results. For instance, Guo et al. \cite{guo2023shadowdiffusion} proposed an unrolling diffusion model that leverages an illumination map and a coarse mask to retrain a diffusion network in producing shadow-free images. Mei et al. \cite{mei2024latent} introduced a diffusion model conditioned on a learned latent feature space that captures the essential characteristics of shadow-free images. Liu et al. \cite{liu2024residual} extended the denoising-diffusion process to image restoration tasks by retraining a diffusion model to estimate noise and residuals simultaneously.  While these diffusion models show promising potential, they often struggle with accurately removing complex shadows due to the limitations imposed by small-scale shadow datasets. To overcome the limitations of datasets and generate clear, realistic texture details, recent researchers have attempted to fine-tune a well-trained diffusion model, such as Stable-diffusion \cite{rombach2022high}, to adapt them for image restoration tasks. Lin et al. \cite{lin2023diffbir} introduced DiffBIR, a pioneering framework for blind image restoration that leverages the capabilities of Stable Diffusion. Building on this foundation, Yu et al. \cite{yu2024scaling} and Wu et al. \cite{wu2024seesr} further expand the multi-modal model to solve real-world image super-resolution problems effectively. Inspired by these breakthroughs in image restoration, our goal is to leverage pre-trained large-scale generative models to shorten the training period and significantly enhance the visual quality of shadow removal by harnessing the generative priors.

\section{Proposed methods}
\begin{figure}
    \centering
    \includegraphics[width=1.0\linewidth]{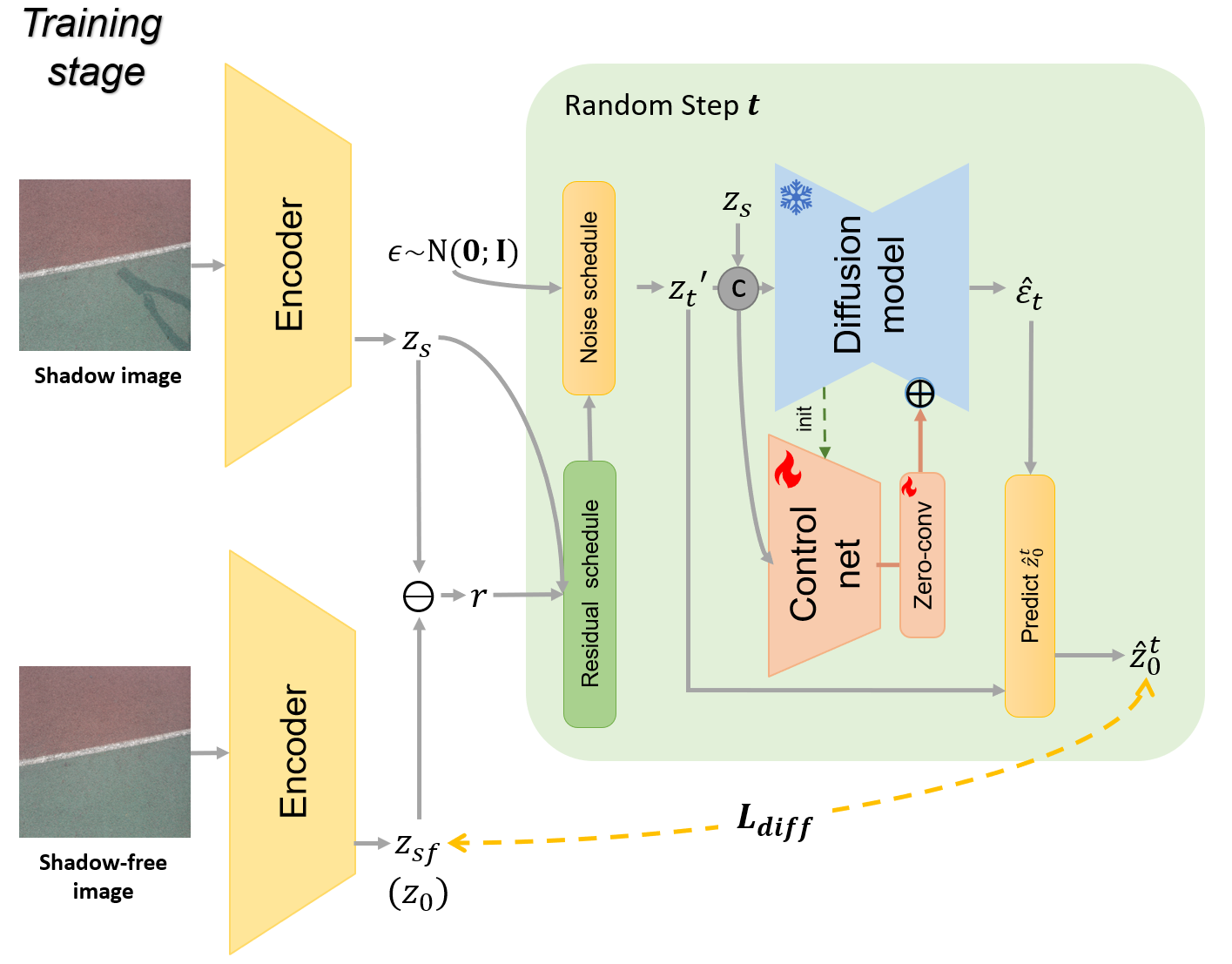}
    \caption{Flowchart of the training phase of the proposed method.}
    \label{fig:train_phase}
\end{figure}

\begin{figure*}
    \centering
    \includegraphics[width=0.98\linewidth]{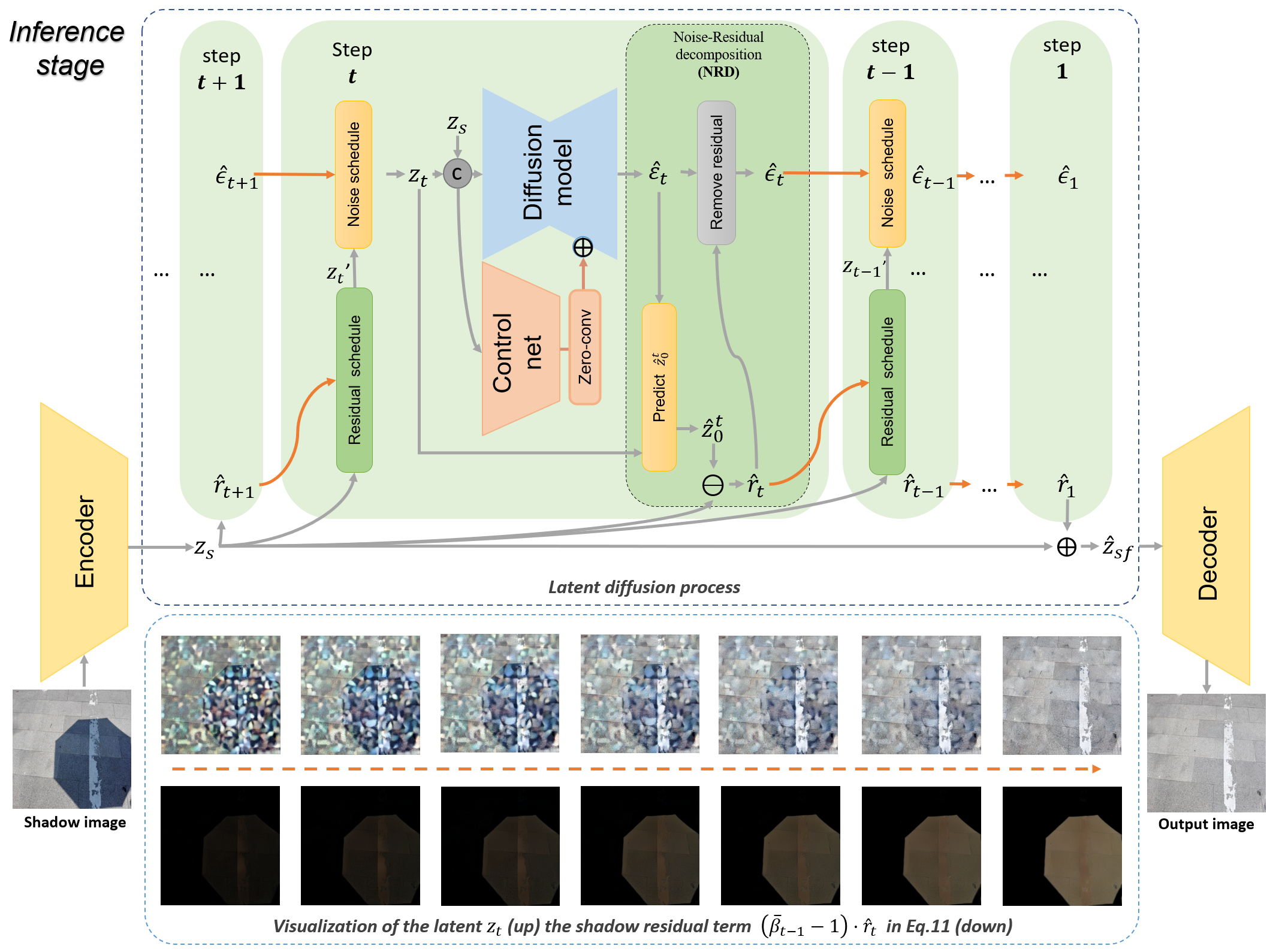}
    \caption{Flowchart of the inference (sampling) phase of the proposed method. The latent $z_t$ and shadow residual term $\left(\bar{\beta}_{t-1}-1\right) \cdot \hat{r}_t$ in Eq. \ref{eq11} are also visualized for better understanding.}
    \label{fig:test_phase}
\end{figure*}
\subsection{Diffusion-based Residual Generation}
\label{section3_1}
Our goal is to leverage a well-trained large-scale diffusion model for mask-free shadow removal. Currently, the large-scale generative models such as Stable Diffusion \cite{rombach2022high}, have made significant advancements in areas like image super-resolution \cite{lin2023diffbir} and inpainting \cite{yang2023paint,TMMinpainting}. These models adhere to specific forward and backward diffusion steps,  achieving the task of generating images from noise. Starting from a clean image $z_0$, the forward diffusion process yields a sequence of images with increasing noise $\left\{z_t\right|t\in[0,T]\}$ via:
\begin{equation}
\label{eq1}
    z_t=\sqrt{\overline{\alpha}_t}\ \cdot z_0+\sqrt{1-\overline{\alpha}_t} \cdot \epsilon, 
\end{equation} 
where $\epsilon$ is the Gaussian noise. ${\alpha}_{t}$ denotes a coefficient associated with the noise schedule, and $\bar{\alpha_t}$ represents the sum of the coefficients from step $0$ to $t$. In the backward diffusion process, the noise is predicted using a trained network and gradually removed for $T$ steps. In Denoising Diffusion Implicit Model (DDIM), as shown in Fig. \ref{fig:DDIM-RDDM}(a), a deterministic sampling strategy is defined as:

\begin{equation}
\hat{z}_0^t=\frac{z_t-\sqrt{1-\bar{\alpha}_t} \cdot \hat{\epsilon}_{t}}{\sqrt{\bar{\alpha}_t}},
\label{eq3}
\end{equation}

\begin{equation}
\quad z_{t-1}=\sqrt{\bar{\alpha}_{t-1}} \cdot \hat{z}_0^t+\sqrt{1-\bar{\alpha}_{t-1}} \cdot \hat{\epsilon}_{t},
\label{eq2}
\end{equation}
where $\hat{\epsilon}_{t}$ denotes the predicting noise at step $t$ and $\hat{z}_0^t$ corresponds to the shadow-free image obtained from it. It is easy to see that Eq. \ref{eq3} and Eq. \ref{eq2} first use the noise estimated by the network to predict the target image. Then, the predicted target image is mixed with the noise at the corresponding scale of the $ t-1 $ step to obtain the input for the network at the $ t-1 $ step.
In order to train a network that estimates noise $\epsilon$, the objective function is set as:

\begin{equation}
\mathcal{L}_{Diff}={E}_{t \sim \operatorname{Uniform}(1, T)} \left[\left\|\epsilon-\hat{\epsilon}_t\right\|_2^2\right],
\label{eq4}
\end{equation}
where $\hat{\epsilon}_t$ denotes estimated noise by the neural network at the timestep $t$.

However, Applying these models for shadow removal presents several significant challenges. First, shadow removal is inherently a deterministic reconstruction task, necessitating the precise reconstruction of image details while effectively eliminating illumination differences between shadowed and non-shadowed areas. This imposes strict demands on diffusion generative models, as each step in the backward process must achieve high precision to avoid error accumulation and preserve essential image details. Second, many large-scale diffusion models, such as Stable Diffusion, employ a VQ-GAN-based encoder to extract latent image features, thereby reducing the computational burden. However, this architecture often sacrifices some original high-frequency details, and the decoder struggles to faithfully reconstruct these missing details, ultimately leading to a decrease in image fidelity.

To address the issues above, we propose a novel approach to fine-tune a pre-trained latent diffusion model (LDM) for image shadow removal. The training process and inference process of the proposed method are respectively shown in Fig. \ref{fig:train_phase} and Fig. \ref{fig:test_phase}. Unlike previous methods, our approach utilizes LDM to generate and refine the shadow residuals between the shadow-free image and the shadowed image rather than regenerating the shadow-free image from pure noise. Specifically, to avoid altering the input-output composition of the pre-trained diffusion model, we introduce a residual schedule to the original diffusion process, facilitating a gradual transition from noisy shadow images to clear, shadow-free images. Let $z_{0} = z_{sf}$ denote the latent representation of a shadow-free image, and $r =z_{s} - z_{0} $  represent the shadow residual between the latent representations of the shadow and shadow-free images. The modified forward diffusion process is as follows:
\begin{equation}
\label{eq5}
\begin{aligned}
    {z}_{t}'&=z_{0}+\bar{\beta_{t}} \cdot r \\
            &=z_{s}+(\bar{\beta_{t}}-1) \cdot r,
\end{aligned}
\end{equation}
\begin{equation}
\label{eq6}
z_t = \sqrt{\overline{\alpha}_t} \cdot {z}_{t}'+\sqrt{1-\overline{\alpha}_t} \cdot \epsilon, 
\end{equation}
where  $\beta$ denotes the coefficient of the shadow residual schedule and $\bar{\beta}_t$ represents the cumulative sum of the $\beta$ coefficients from step $0$ to step $t$. Following the framework of residual denoising diffusion models (RDDM), $\beta$ is linearly increased from $0$ to $1$ over time steps from $0$ to $T$. Clearly, Eq.\ref{eq5} corresponds to an interpolation operation between the shadow and shadow-free image, whereas Eq. \ref{eq6} outlines the noise and image mixing strategy in Eq.\ref{eq1}. During the network training phase, as shown in Fig. \ref{fig:train_phase}, we utilize the noise predicted by the network to estimate a clean shadow-free latent representation. The loss function is set to:
\begin{equation}
L_{{diff}}:=E_{t \sim \operatorname{Uniform}(1, T)}\left[\left\|\hat{z}_{0}^{t}-{z}_{0}\right\|_2^2\right] {, }
\label{eq7}
\end{equation}
\begin{equation}
    \hat{z}_{0}^{t} = \frac{{z}_t-\sqrt{1-\bar{\alpha}_t} \cdot \hat{{\varepsilon}}_t}{\sqrt{\bar{\alpha}_t}} {,}
    \label{eq8}
\end{equation}
where $\hat{z}_{0}^{t}$ represents the estimated noise at step $t$, and $\hat{{\varepsilon}}_t$ denotes the predicted output by the network at that step. the proposed network is developed based on ControlNet \cite{zhang2023adding}, a neural network plugin that guides the generation process by adjusting the intermediate features of a fixed diffusion model. Given our integration of the shadow residual into the forward diffusion process and adhering to the structural framework in Eq. \ref{eq3} for predicting the clear image, the $\hat{{\varepsilon}}_t$ transcends being merely an estimation of the applied noise $\epsilon$ in Eq. \ref{eq6} and additionally encompasses an estimation of the residual. Assuming $\hat{z}_{0}^{t}={z}_{0}$, substituting Eq. \ref{eq5} and Eq. \ref{eq6} into Eq. \ref{eq8} and rearranging, one can obtain:
\begin{equation}
\label{eq9}
    \begin{aligned}
    \hat{\varepsilon}_t &=\frac{z_{t}-\sqrt{\bar{\alpha}_{t}} \cdot \hat{z}_{0}^{t}}{\sqrt{1-\bar{\alpha}_{t}}}  \\ 
    &=\frac{z_{t}- \sqrt{{\bar{\alpha}}_{t}} \cdot {z}^{'}_{t}}{\sqrt{1-\bar{\alpha}_{t}}}+\frac{\bar{\beta}_{t} \sqrt{\bar{\alpha}_{t}} \cdot r}{\sqrt{1-\bar{\alpha}_{t}}} \\
    &=\epsilon + \frac{\bar{\beta}_{t} \sqrt{\bar{\alpha}_{t}} \cdot r}{\sqrt{1-\bar{\alpha}_{t}}}.
    \end{aligned}
\end{equation}

It is evident that $\hat{\varepsilon}_t$ inherently encapsulates the shadow component. Consequently, the network has transitioned from solely estimating noise from a noisy latent to concurrently estimating both the noise and the residuals. Due to the large-scale models being well-trained for noise estimation, the newly introduced control network can focus on estimating the residuals. In this work, a pre-trained inpainting model \cite{yang2023paint} is employed as the backbone model. By incorporating the latent representation of a shadow image and an all-one mask as auxiliary inputs for the noisy image (indicating that no inpainting is applied), the structure of the input image can be well preserved, thereby enhancing the detail fidelity of the proposed framework. A detailed comparison between the use of an inpainting model and a text-to-image generation model (Stable Diffusion \cite{rombach2022high}) will be provided in Section \ref{section4_3}.



To leverage the framework presented in this paper for inferring shadow-free images, a Noise-Residual Decomposition (NRD) approach is initially introduced to decompose the output of the diffusion network into residual and noise components. Subsequently, the shadow residual and noise schedule are integrated into the shadow image latent to produce the input for the network at the next time step. Specifically, the estimated shadow residual map $\hat r_t$ is first extracted using:
 
 \begin{equation}
\hat r_t = {z}_\text{s}-\hat{z}_{0}^{t},
\label{eq10} 
\end{equation}
where $\hat{z}_{0}^{T}$ is initialized to be $z_s$ at the beginning sampling step $T$. The obtained shadow residual is subsequently added to the latent representation of the shadow image. Referring to diffusion forward process in Eq. \ref{eq5}, one can obtain:
 \begin{equation}
{z}_{t-1}'= z_s + \left(\bar{\beta}_{t-1}-1\right) \cdot \hat{r}_t.
\label{eq11} 
\end{equation}
Referring to Eq. \ref{eq9}, the noise $\hat{\epsilon}_t$ present in the network estimation $\hat{\varepsilon}_t$ can be derived through:
\begin{equation}
   \hat{\epsilon}_t =\hat{{\varepsilon}}_t - \frac{\bar{\beta}_{t} \sqrt{\bar{\alpha}_{t}} \cdot \hat{r}_t}{\sqrt{1-\bar{\alpha}_{t}}}.
\label{eq12} 
\end{equation}
By incorporating the noise schedule akin to Eq.6, the final sampling formula for the backward diffusion process can be derived as:
\begin{equation}
   z_{t-1} = \sqrt{\overline{\alpha}_{t-1}} \cdot {z}_{t-1}'+\sqrt{1-\overline{\alpha}_{t-1}} \cdot \hat{\epsilon}_t.
   \label{eq13} 
\end{equation}

 Eq. \ref{eq13} describes a non-Markovian process wherein the inference of $z_{t-1}$ depends on both preceding state $z_t$ and shadow image latent $z_s$. This aligns with DDIM and enables the proposed method to employ interval sampling similarly to reduce the number of sampling steps. As a result, the proposed method does not require alterations to the original noise strategy. At the end of sampling, we directly add the estimated shadow residual $\hat{r}_1$ to $z_s$ to yield a shadow-free latent $\hat{z}_{sf}$ and subsequently decode it to produce the final shadow removal result. 


Fig. \ref{fig:DDIM-RDDM} provides a detailed comparison of the proposed method with previous diffusion models DDIM \cite{song2020denoising} and RDDM \cite{liu2024residual}. As shown in Fig. \ref{fig:DDIM-RDDM}(a), the DDIM employs a sole denoising strategy, which, while effective for generating a shadow-free image from pure noise, may lead to the loss or distortion of detailed information. 
Building upon this, RDDM (Fig. \ref{fig:DDIM-RDDM}(b)) further introduces residual estimation as an auxiliary task, paralleling the process of gradually removing noise and residuals from the image (or latent representation). However, due to the change in noise schedule and the need for additional branches or networks to estimate residuals, it is necessary to retrain the backbone model, thus missing the full use of the pre-trained large model. It is worth noting that RDDM also provides a parameter conversion strategy between DDIM and RDDM to facilitate the application of DDIM noise strategy on the RDDM model. Nevertheless, this strategy assumes $z_s=0$, implying the regeneration of $z_0$ from scratch without associating to the shadow image latent. This contrasts the high-fidelity objectives essential for the shadow removal task.

In contrast, the proposed method does not alter the noise schedule of the backbone model but incorporates a shadow residual schedule on top of it. The aim is to leverage the generative priors of the well-trained diffusion model to produce and refine the shadow residuals with minimal modifications to the model itself. As shown in Fig. \ref{fig:DDIM-RDDM}(c), a control net initialized by the pre-trained diffusion model is added without introducing additional branches with random initializations, thereby enhancing the stability of the training process. In addition, by establishing dense connections with the shadow-affected image at each diffusion step to guide the generative trajectory, the proposed method can preserve the original content and minimize unintended alterations.

\subsection{Cross-timestep Self-enhanced Training}
\begin{figure}
    \centering
    \includegraphics[width=0.9\linewidth]{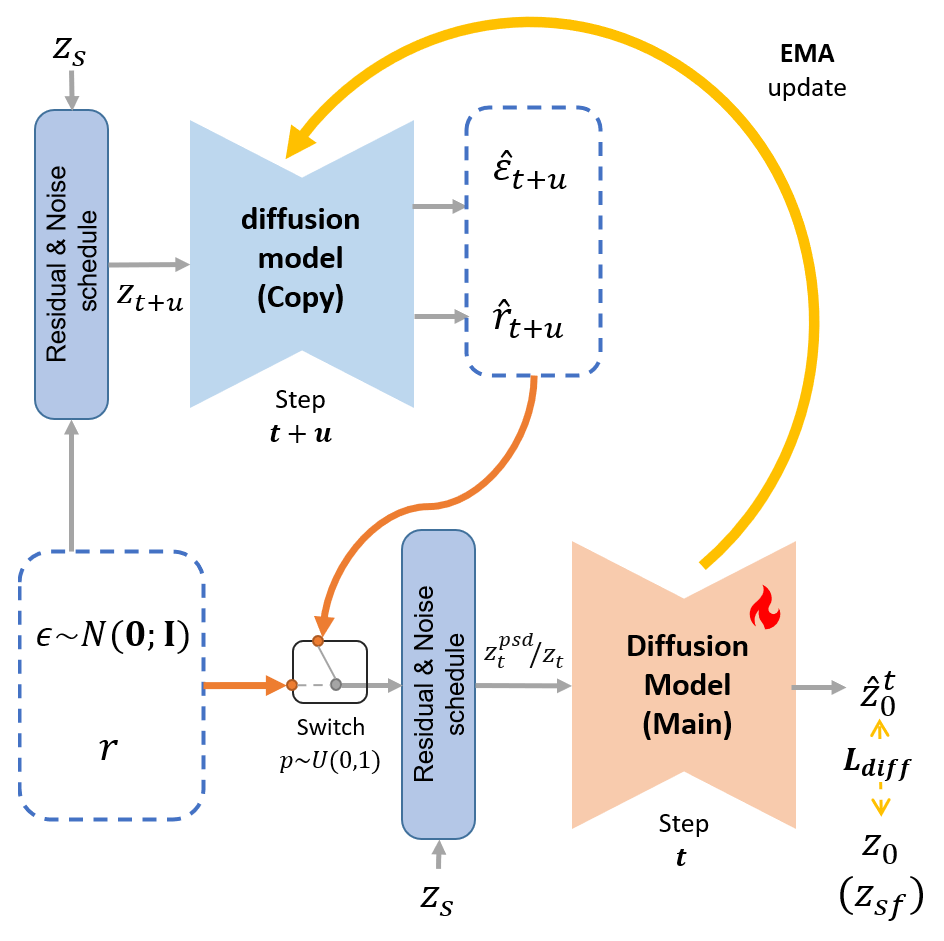}
    \caption{The schematic illustration of our training strategy.}
    \label{fig:EMA}
\end{figure}

In the training phase of diffusion models, each time step is trained independently with real data combined with random perturbations, often neglecting the consideration of discrepancies between the real data and the network's output. In the case of deterministic DDIM sampling in Eq. \ref{eq3} and Eq. \ref{eq2}, the input to the network at time step $(t-1)$ relies on the noise map $z_t$ and the noise estimate $\hat{\epsilon}_t$ from the previous time step $t$. Consequently, when the predictive accuracy of the network is inadequate, it may lead to the accumulation of errors, which can cause deviations in the generated trajectory and alterations in image details.

To tackle this issue, we aim for each step of the model to be able to rectify errors from the preceding steps. To achieve this, a cross-timestep self-enhancing training mechanism is proposed. As shown in Fig. \ref{fig:EMA}, a copy of the network is created at the current state, which is initially detached from the main network. The weights of this copy are then updated by means of the exponential moving average (EMA) \cite{he2020momentum}. The EMA simply updates the weights of the copy network by taking a linear combination of the weights from the main network and the copy network:
\begin{equation}
  w_{\text{copy}} = \eta \cdot w_{\text{main}} + (1 - \eta) \cdot w_{\text{copy}}
\label{eq14}  
\end{equation}
where $ w_{\text{copy}}$ and $w_{\text{main}}$ denote the weights of the copy and main networks, respectively, and $\eta$ represents the smoothing factor that is empirically set to $0.999$. 

Let $t+u$ represent a randomly selected prior timestep, where $u=Rand(1,50)$ denotes a random integer chosen uniformly from 1 to 50. When the probability threshold $p$ is less than $P$, we employ Eq. \ref{eq5} and Eq. \ref{eq6} to blend the genuine shadow image latent $z_s$, shadow latent residuals $r$, and random noise $\epsilon$ to create the noise latent $z_{t+u}$. Subsequently, Eq. \ref{eq13} is applied for backward sampling to derive the pseudo-input {$z^{psd}_t$} for the current timestep $t$, which is then fed into the main network. The loss function, as delineated in Eq. \ref{eq7}, is utilized to compel the primary network to estimate an accurate shadow-free image, thereby correcting the errors present in the pseudo-input. In the opposite scenario where $p$ exceeds $P$, we straightforwardly apply Eq. \ref{eq5} and Eq. \ref{eq6} to apply the real shadow image latent, shadow latent residuals, and random noise to prepare the input of network at timestep $t$ ($z_t$) for training purposes. Empirically, the hyperparameter $P$ is set to 0.2.

This training methodology not only permits the network to rectify the generation process during the backward diffusion phase but also serves as an effective form of data augmentation. By leveraging a network copy to produce additional training data, the training dataset is expanded, consequently enhancing the network's capabilities in shadow prediction and removal. In practice, since the backbone model parameters are fixed and the replica network shares the same backbone as the primary network, one can simply duplicate the control network modules to reduce memory costs significantly.

\subsection{Detail-preserving Image Reconstruction}
\begin{figure*}
    \centering
    \includegraphics[width=0.6\linewidth]{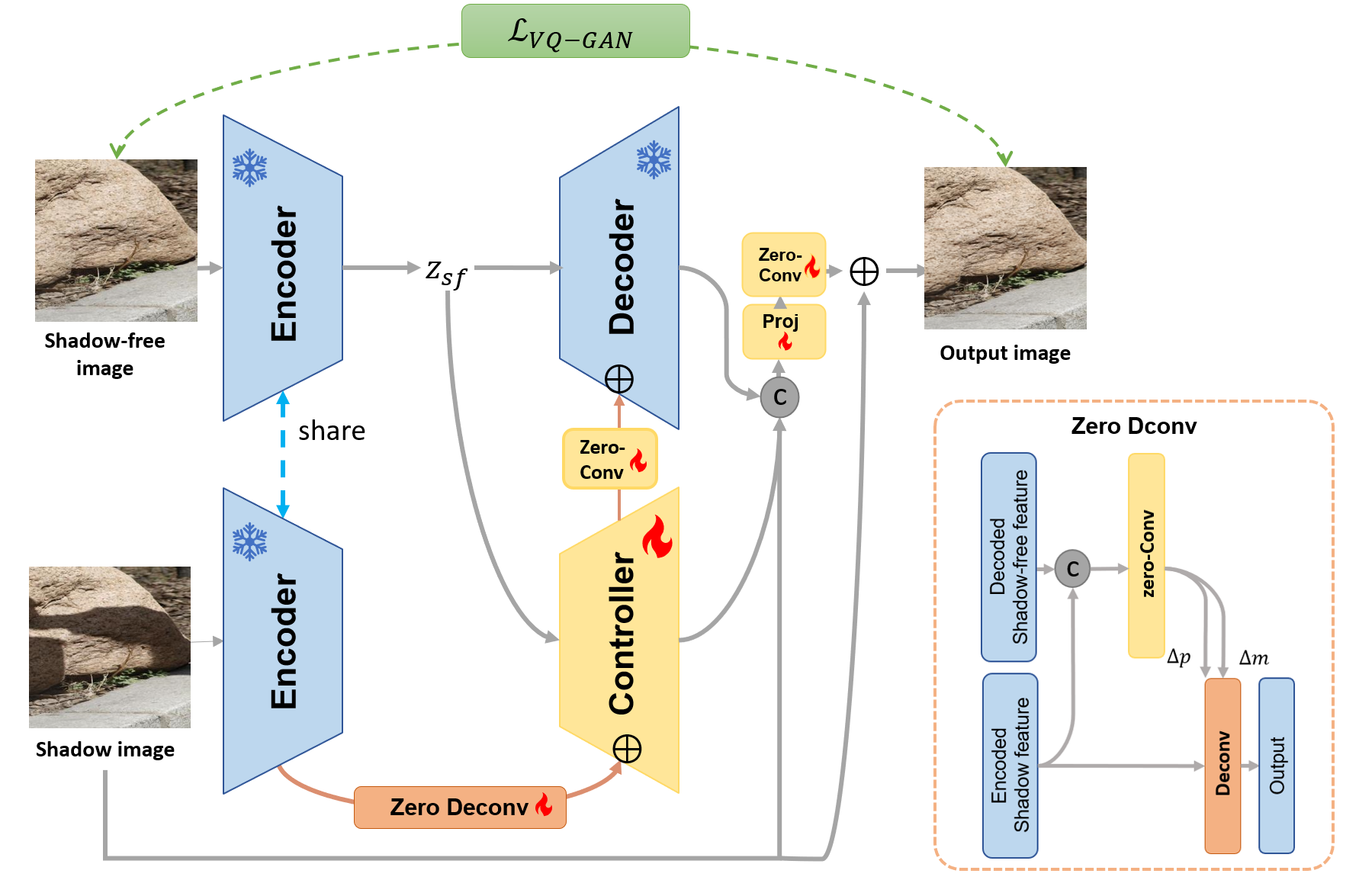}
    \caption{The structure of the proposed detail-preserving decoder.}
    \label{fig:decoder}
\end{figure*}

To minimize computational overhead, Existing large-scale generative models often employ a pre-trained encoder, such as VQ-GAN \cite{esser2021taming}, to shrink the spatial dimensions of images to the low-resolution latent space before the diffusion process. Despite the encoder and decoder of the VQ-GAN being trained in pairs, they merely share information at the smallest scale, leading to the loss of original image details. During testing, we observed that images reconstructed by the original decoder often exhibited curling in texture details. For instance, the characters in the text region were often distorted into unrecognizable symbols. These issues significantly impacted the visual quality of the reconstructed images and had consequences for downstream tasks.
 
 To improve missing details caused by the encoder-decoder structure, Li et al. \cite{zhu2023designing} fine-tuned the decoder to prevent alterations in content outside the mask for the image inpainting task. They incorporated a conditional encoder into the decoder, reconstructing the image by combining features from the conditional encoder with the original features at various levels, guided by the mask. Unfortunately, this structure faced difficulties when directly applied to shadow removal tasks. Obtaining an accurate mask for shadows, especially for soft shadows with indistinct boundaries, is often challenging. Relying on an accurate mask for reconstruction is not flexible and significantly limits the application scenarios of the decoder. Furthermore, training the decoder may compromise its inherent reconstruction capabilities, unnecessarily increasing training costs. 


Inspired by ControlNet \cite{zhang2023adding}, we propose a detail-preserving decoder architecture to address the aforementioned issues. Specifically, we freeze the original encoder and decoder of the VQ-GAN and introduce a controller to regulate the reconstruction process. Similar to ControlNet, the controller is initialized with the weights of the original decoder and receives $z_{sf}$ as input features. To fully exploit the multi-scale information present in the shadow image, we encode the shadow image information from the encoder and establish skip connections between the encoder and the controller. To address potential misalignments in the latent representations between shadow-free images and shadowed images, the zero-initialization strategy is applied to deformable convolutions \cite{zhu2019deformable}, termed as Zero-Deconv, which is then added into the skip connections. The structure of Zero-Deconv is shown in Fig. \ref{fig:decoder}, which can be calculated as:
\begin{equation}
  \text{Zero-DeConv}(F_{s})=\sum_{k=1}^K w_k \cdot F_{s}\left(p+p_k+\Delta p_k\right) \cdot \Delta s_k
  \label{eq15} 
\end{equation}
where $F_s$ denotes the encoded shadow image feature at one of the scales from the encoder, $k$ is the spatial size of kernel, $p$ represents a specific coordinate position within the feature map, and $\Delta p$ and $\Delta s$ correspond to the learnable offset and modulation scalar, respectively. As depicted in Fig. \ref{fig:decoder}, the encoded shadow features are concatenated with the shadow-free features from the encoder and fed the combined features into a zero-initialized convolutional (Zero-Conv) layer to learn an initial value of ($\Delta p$,$\Delta s$)=(0,0). This initialization ensures that the added skip connections do not introduce any disruptive effects to the controller in their initial state, allowing for a smooth integration of the features during the initial training phases. To refine the output of the decoder in the image domain and generate an image residual, we further concatenate the outputs of the decoder and the controller with the shadow image and feed it into a project module followed by a Zero-Conv layer. This residual is then added to the shadow image to obtain the final result. This new decoder structure is trained using the loss function from the standard VQ-GAN ($\mathcal{L}_{VQ-GAN}$). In the inference phase, $z_{sf}$ is estimated through the backward diffusion process. The proposed detail-preserving decoder simultaneously incorporates the latent representations of both shadow-free and shadow images into the controller. This setup enables the controller to implicitly identify the shadow regions, thereby facilitating a shadow-aware reconstruction process. By harnessing the multi-scale information extracted from the shadow image by the encoder, the proposed approach can generate high-fidelity outputs and significantly preserve original image details during the feature encoding phase.

\section{Experiments}
\subsection{Implementation Details}
The proposed algorithm is implemented based on a pre-trained Paint-By-Example \cite{yang2023paint} model, wherein the conditional input is encoded by leveraging the image encoder of CLIP \cite{radford2021learning}. In this paper, we utilize the shadowed image itself as the condition input. During training, the AdamW optimizer is applied with momentum parameters set to 0.9 and 0.999, along with a weight decay factor of 0.01. The learning rate begins at 5e-5 and is decremented to 1e-6 using a cosine annealing technique. Images are preliminarily resized to dimensions between $256\times256$ and $288\times288$ pixels, followed by random cropping to extract $256\times256$ pixel training patches.

\subsection{Datasets}
The proposed framework is trained and evaluated using two prominent shadow removal benchmark datasets: ISTD+ \cite{le2021physics}, and SRD \cite{RN7}. The ISTD+ dataset consists of 1330 shadow image triplets, including the shadow image, mask, and shadow-free image, for training, and 540 triplets for testing. The SRD dataset provides 2680 image pairs for training and 408 for testing, without the inclusion of accompanying image masks.

\subsection{Evaluation Metrics}
Following prior research \cite{fu2021auto,li2023leveraging}, we implemented a mask-based image decomposition approach to analyze different aspects of our results. Specifically, we evaluated the peak signal-to-noise ratio (PSNR) and Structural Similarity Index (SSIM) for shadowed and non-shadowed regions, as well as the overall image. For the evaluation of the SRD dataset, we utilize publicly available shadow masks from the method \cite{cun2020towards} for assessment. Additionally, we utilized the Learned Perceptual Image Patch Similarity (LPIPS) \cite{RN33} and Fréchet Inception Distance (FID) \cite{heusel2017gans}  to measure perceptual differences between the reconstructed and ground truth images.

\subsection{Comparison with SOTA methods}
We compared our approach with state-of-the-art mask-free (inference-stage) shadow removal methods, including STC-GAN \cite{wang2018stacked}, DSC \cite{hu2018direction}, Mask-ShadowGAN \cite{hu2019mask}, DHAN \cite{cun2020towards}, LG Net \cite{liu2021shadow}, DC \cite{jin2021dc}, DMTN \cite{liu2023decoupled}, TBRN \cite{liu2023shadow}, DA-SDE \cite{luo2023controlling} and DeS3 \cite{jin2024des3}. Consistent with the comparative methods in previous works, we conducted tests and evaluations on images of a uniform size of $256\times256$ pixels. When feasible, we utilized the codes and weights provided by the authors for the compared methods. In cases where these resources were not accessible, we relied on the outcomes reported by a benchmark test provided by \cite{hu2024unveiling} to ensure a fair and comprehensive comparison.

\begin{figure*}
    \centering
    \includegraphics[width=0.95\linewidth]{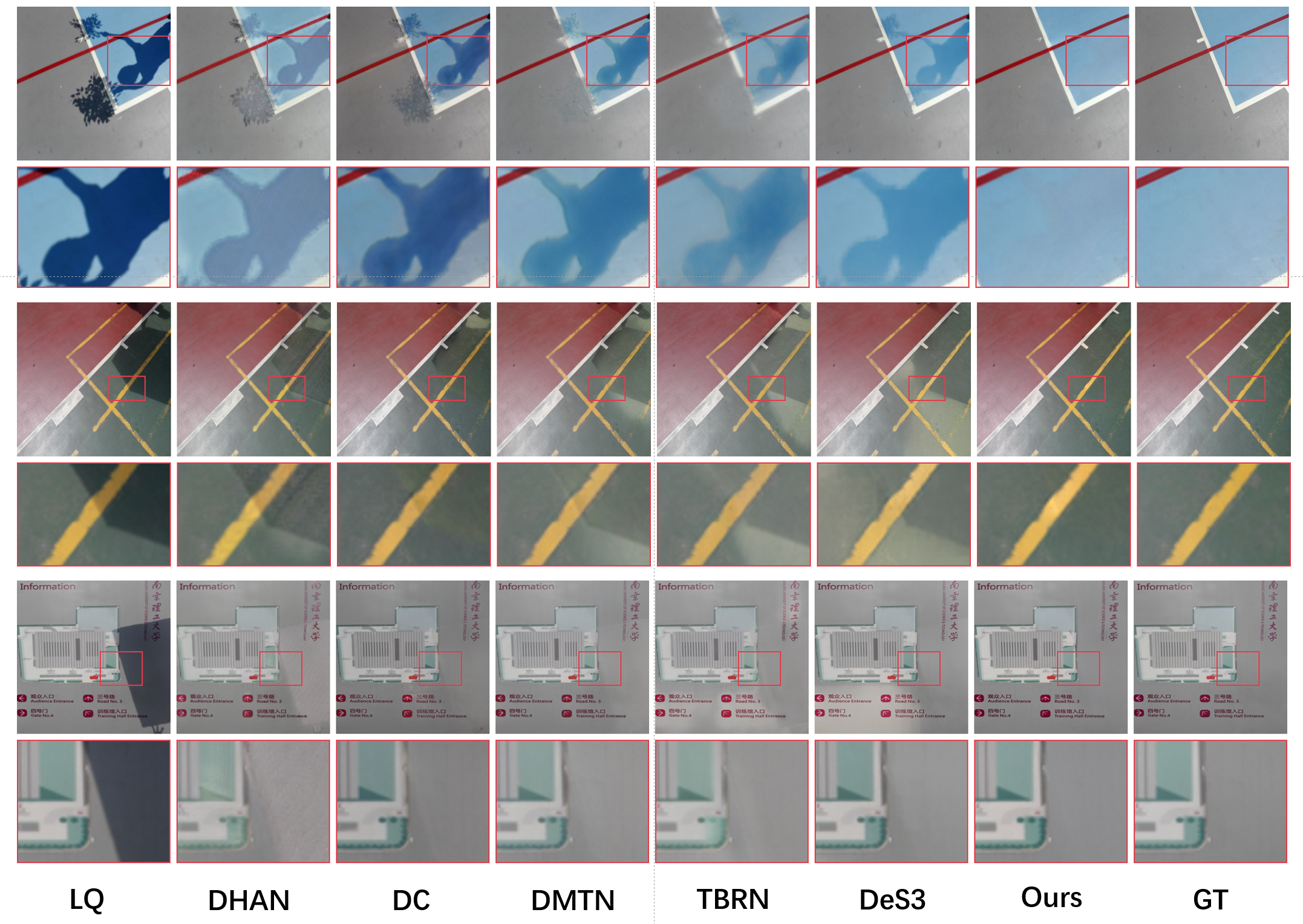}
    \caption{Visual comparison of different methods without input masks on the ISTD+ dataset, with enlarged views of shadow edges for clearer contrast. } 
    \label{fig:ISTD+}
\end{figure*}

\begin{figure*}
    \centering
    \includegraphics[width=0.95\linewidth]{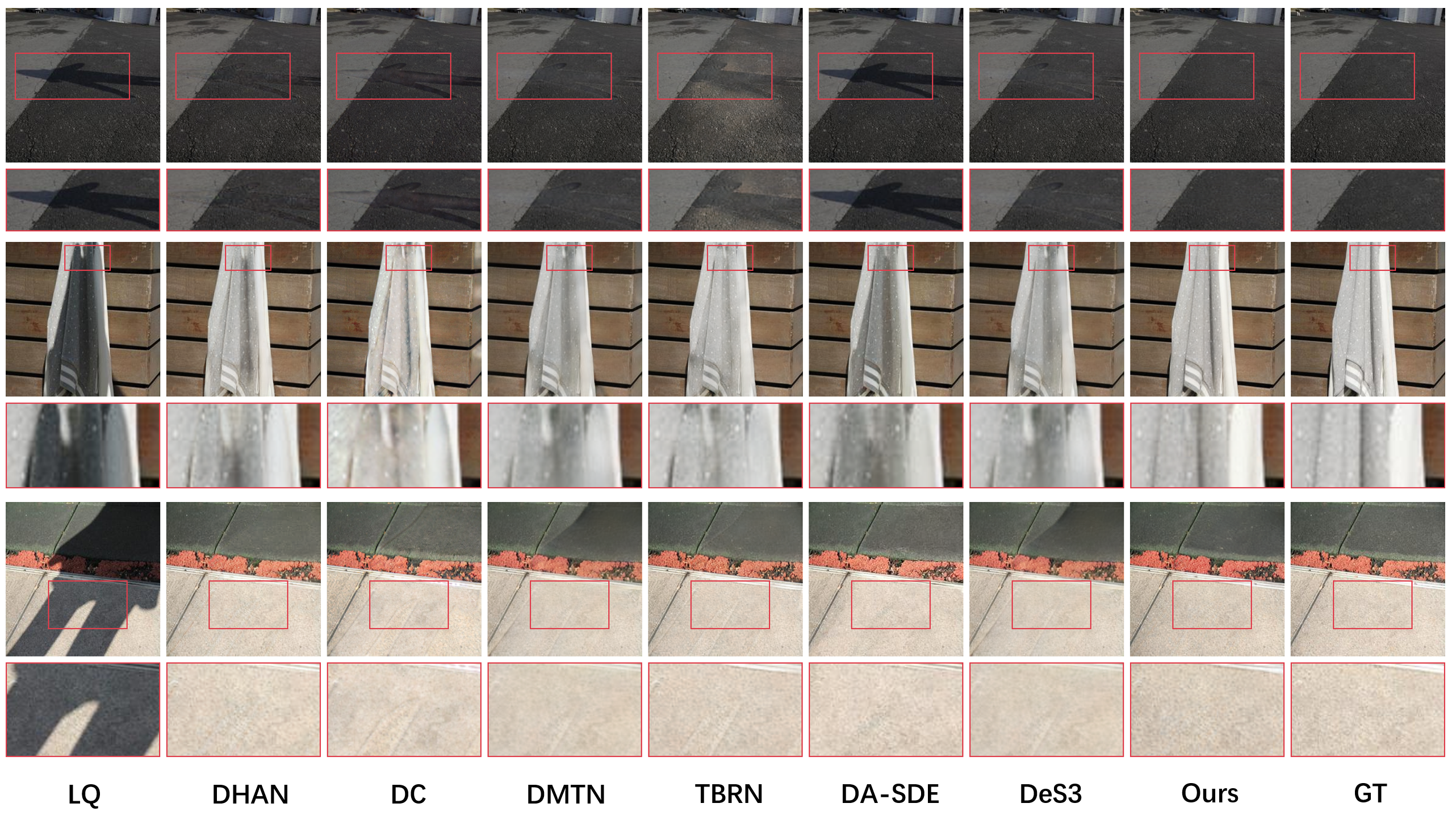}
    \caption{Visual comparison of different methods without input masks on the SRD dataset, with enlarged views of shadow edges for clearer contrast.}
    \label{fig:srd}
\end{figure*}

\textbf{Qualitative Evaluation.} Fig. \ref{fig:ISTD+} and Fig. \ref{fig:srd} compare the shadow removal outputs of our proposed method with SOTA methods on challenging scenes from the ISTD+ and SRD datasets, respectively. In the ISTD+ dataset, the test set contains a wide range of background scenes that differ from those in the training set. During testing, we observed that while previous methods could somewhat reduce the illumination difference between shadow and non-shadow regions, they still struggled with removing complex shadows. As shown in the first row of Fig. \ref{fig:ISTD+}, previous approaches often fail to remove human shadows. In the third and fifth rows, these methods could mistakenly identify dark areas in the image as shadows, leading to inaccurate shadow removal. In the SRD dataset, previous methods frequently produced boundary artifacts at shadow edges, resulting in an unnatural appearance of results. In contrast, beneficial from the generative priors from pre-trained large-scale generative models, our proposed method significantly reduces perceptual differences between shadowed and non-shadowed areas and effectively suppresses boundary artifacts. Notably, the enlarged detailed images demonstrate that our method effectively preserves the content and details of the shadowed image, validating the effectiveness of the proposed fidelity strategies.

\begin{table*}[!t]
\label{ISTD+}
\renewcommand\arraystretch{1.3}
\caption{Quantitative evaluation of various methods without input masks on the ISTD+ Dataset. Bold text indicates the best score, while underlined text represents the second-best score.}
\centering
\resizebox{350pt}{!}{\begin{tabular}{ccccccccc}
\hline
                 \textbf{Method} & \textbf{PSNR}                & \textbf{PSNR-NS}              & \textbf{PSNR-S}             & \textbf{SSIM}                & \textbf{SSIM-NS}              & \textbf{SSIM-S}             & \textbf{LPIPS↓}  & \textbf{FID↓}             \\
\hline
\textbf{STC-GAN\cite{wang2018stacked}} & 29.95                        & 33.92                        & 34.21                        & 0.937                        & 0.963                        & 0.982                        & 0.0717  &51.982                     \\
\textbf{DSC\cite{hu2018direction}}    & 29.47                        & 31.66                        & 35.40 & 0.930  &                       0.931                       & 0.982                       & 0.0854 &51.612 \\
\textbf{Mask-ShadowGAN\cite{hu2019mask}}    & 28.48                        & 33.09                        & 31.70 & 0.938  &                       0.971                       & 0.980                       & 0.0754 & 55.174\\
\textbf{LG Net\cite{liu2021shadow}}    & 28.28                        & 33.45                       & 30.85 & 0.937  &                       0.974                       & 0.979                       & 0.0880 & 71.586 \\ 
\textbf{DHAN\cite{cun2020towards}}    & 25.65                        & 27.14                        & 32.91 & 0.955  &                       0.970                       & 0.987                       & 0.0798 & 28.986 \\
\textbf{DC Net\cite{jin2021dc}}  & 28.79 & 33.57 & 32.01                        & 0.931                        & 0.967                        & 0.976                        & 0.0961       &60.205                 \\
\textbf{DMTN\cite{liu2023decoupled}}    & 31.80                        & \underline{35.79}                        & 35.73 & 0.963  &                       \textbf{0.978}                       & \underline{0.990}                       & 0.0351 & 24.021\\
\textbf{TBR Net\cite{liu2023shadow}}    & \underline{31.89}                        & 35.52                        & 36.33 & \underline{0.963}  &                       \underline{0.976} & \underline{0.990} & \underline{0.0332} & 22.057\\
\textbf{DeS3\cite{jin2024des3}}    & 31.37                        & 34.69                        & 36.49 & 0.957  &                       0.972 & 0.989 & 0.0350 & \underline{21.724}\\
\textbf{Ours} & \textbf{33.38 }                       & \textbf{36.41 }                       & \textbf{37.92}                       & \textbf{0.965}                        & \textbf{0.978}                        & \textbf{0.991}                        & \textbf{0.0287} &  \textbf{17.723}     
\\
\hline
\end{tabular}}
\label{table_istd+}
\end{table*}

\begin{table*}[!t]
\renewcommand\arraystretch{1.3}
\caption{Quantitative evaluation of various methods without input masks on the SRD Dataset. Bold text indicates the best score, while underlined text represents the second-best score.}
\centering
\resizebox{350pt}{!}{\begin{tabular}{ccccccccc}
\hline
                 \textbf{Method} & \textbf{PSNR}                & \textbf{PSNR-NS}              & \textbf{PSNR-S}             & \textbf{SSIM}                & \textbf{SSIM-NS}              & \textbf{SSIM-S}             & \textbf{LPIPS↓}    & \textbf{FID↓}            \\
\hline
\textbf{STC-GAN \cite{wang2018stacked}} & 29.01                        & 33.49                        & 31.79                        & 0.923                        & 0.975                        & 0.966                        & 0.0975   &33.626                    \\
\textbf{DSC \cite{hu2018direction}}    & 27.47                        & 31.15                        & 30.85 & 0.890  &                       0.956                       & 0.963                       & 0.1168 & 39.305\\
\textbf{Mask-ShadowGAN \cite{hu2019mask}}    & 28.08                        & 32.89                       & 30.65 & 0.924  &                       0.977                       & 0.965                       & 0.0907 & 41.068 \\
\textbf{DHAN \cite{cun2020towards}}     & 29.88                        & \underline{34.90}                        & 32.30 & \underline{0.940}  &                       \textbf{0.984}                      & 0.971                       & \underline{0.0645} & \underline{28.534}\\
\textbf{DC Net \cite{jin2021dc}}  & 29.27 & 33.19 & 32.50                        & 0.922                        & 0.971                        & 0.970                        & 0.0961       & 57.403                 \\
\textbf{DMTN \cite{liu2023decoupled}}    & 27.98                        & 33.59                        & 30.13 & 0.929  &                       \underline{0.981}                       & 0.964                       & 0.0785 & 39.404\\
\textbf{TBR Net \cite{liu2023shadow}}    & 29.65                        & 34.47                        & 32.16 & 0.938  &                       0.980 & 0.968 & 0.0676 & 32.321\\
\textbf{DA-SDE \cite{luo2023controlling}}    & 26.92                        & 32.38                        & 29.29 & 0.906  &                       0.974 & 0.953 & 0.0879 & 32.084\\
\textbf{DeS3 \cite{jin2024des3}}    & \textbf{30.57}                        & 33.84                        & \textbf{34.55} & 0.933  &                       0.970 & \textbf{0.977} & 0.0654 & 39.409\\
\textbf{Ours}    & \underline{30.45} & \textbf{35.06} & \underline{33.17}                        & \textbf{0.944} & \textbf{0.984} & \underline{0.973} & \textbf{0.0537}  & \textbf{22.433}\\
\hline
\end{tabular}}
\label{table_srd}
\end{table*}

\begin{table*}[!t]
\renewcommand\arraystretch{1.3}
\caption{Quantitative evaluation of various methods using input masks on the ISTD+ Dataset. Bold text indicates the best score, while underlined text represents the second-best score.}
\centering
\resizebox{350pt}{!}{
\begin{tabular}{ccccccccc}
\hline
                 \textbf{Method} & \textbf{PSNR}                & \textbf{PSNR-NS}              & \textbf{PSNR-S}             & \textbf{SSIM}                & \textbf{SSIM-NS}              & \textbf{SSIM-S}             & \textbf{LPIPS↓}  &\textbf{FID↓}               \\
\hline
\textbf{EPF Net\cite{fu2021auto}} & 29.44                        & 31.15                        & 36.04                        & 0.861                        & 0.892                        & 0.978                        & 0.1119 &50.068 \\
\textbf{SP+M+I Net\cite{le2021physics}} & 33.81                        & 37.28                        & 37.63                        & 0.968                        & 0.983                        & \underline{0.990}                        & 0.0304 &21.662 \\
\textbf{G2R-Net\cite{liu2021shadow}}    & 26.19                        & 32.35                        & 28.51 & 0.898  &                       0.958                      & 0.958                       & 0.1318 & 120.51 \\ 
\textbf{SG Net\cite{wan2022style}} & 33.47                        & 37.16                        & 36.84                        & 0.963                        & \underline{0.983}                        & 0.988                        & 0.0368 &27.889 \\
\textbf{BM Net\cite{zhu2022bijective}} & 33.46                        & 37.43                        & 36.57                        & 0.963                        & 0.982                        & 0.988                        & 0.0440 &33.331\\
\textbf{Inpainting Net\cite{li2023leveraging}} & 34.13                        & 37.59                        & 38.09                        & 0.967                        & 0.980                        & \underline{0.990}                        & \underline{0.0298} & \underline{22.646} \\  
\textbf{ShadowFormer\cite{guo2023shadowformer}} & 34.22                        & 37.70                        & 37.86                        & 0.965                        & 0.982                        & \underline{0.990}                        & 0.0366    &  25.056               \\
\textbf{Latent shadow diffusion\cite{mei2024latent}}    & \underline{34.72}                        & \underline{37.76}                        & \textbf{39.17} & \textbf{0.973}  &                       \textbf{0.984} & \textbf{0.992} & 0.0597 & 47.841 \\
\textbf{HomoFormer\cite{xiao2024homoformer}}    & 34.15                        & 36.02                        & 37.85 & 0.942  &                       0.958 & 0.989 & 0.0391 &23.056 \\
\textbf{Ours-w/ Mask} & \textbf{34.73}                      & \textbf{37.89 }                     &  \underline{38.89}                       & \underline{0.970}                        & \underline{0.983}                        & \textbf{0.992}                        & \textbf{0.0228} & \textbf{14.388}                     \\
\hline
\end{tabular}
}
\label{table_ISTD+_mask}
\end{table*}

\textbf{Quantitative Evaluation.} To further demonstrate the content fidelity of the proposed method, commonly used PSNR, SSIM, and LPIPS are adopted. The quantitative test results on the ISTD+ and SRD datasets are presented in Tables \ref{table_istd+} and \ref{table_srd}. In our comparative analysis, our method has achieved the best LPIPS and FID scores on both the ISTD+ and SRD datasets, aligning with the qualitative assessment conclusions. This indicates that our approach, based on a pre-trained diffusion model, effectively removes image shadows and yields the best visual performance. As previously mentioned, large-scale generative models excel at producing perceptually realistic and clear details by capturing high-level semantics but often struggle to align at the pixel level, resulting in suboptimal performance on fidelity metrics. However, due to the high-fidelity design, the proposed method can achieve comparable or superior PSNR and SSIM metrics to current SOTA non-generative methods. In summary, the proposed method can achieve high-fidelity and high-performance shadow removal, surpassing existing methods in terms of shadow removal capabilities.

\subsection{Extend Analysis on Mask-available Shadow Removal}
\begin{figure*}
    \centering
    \includegraphics[width=0.95\linewidth]{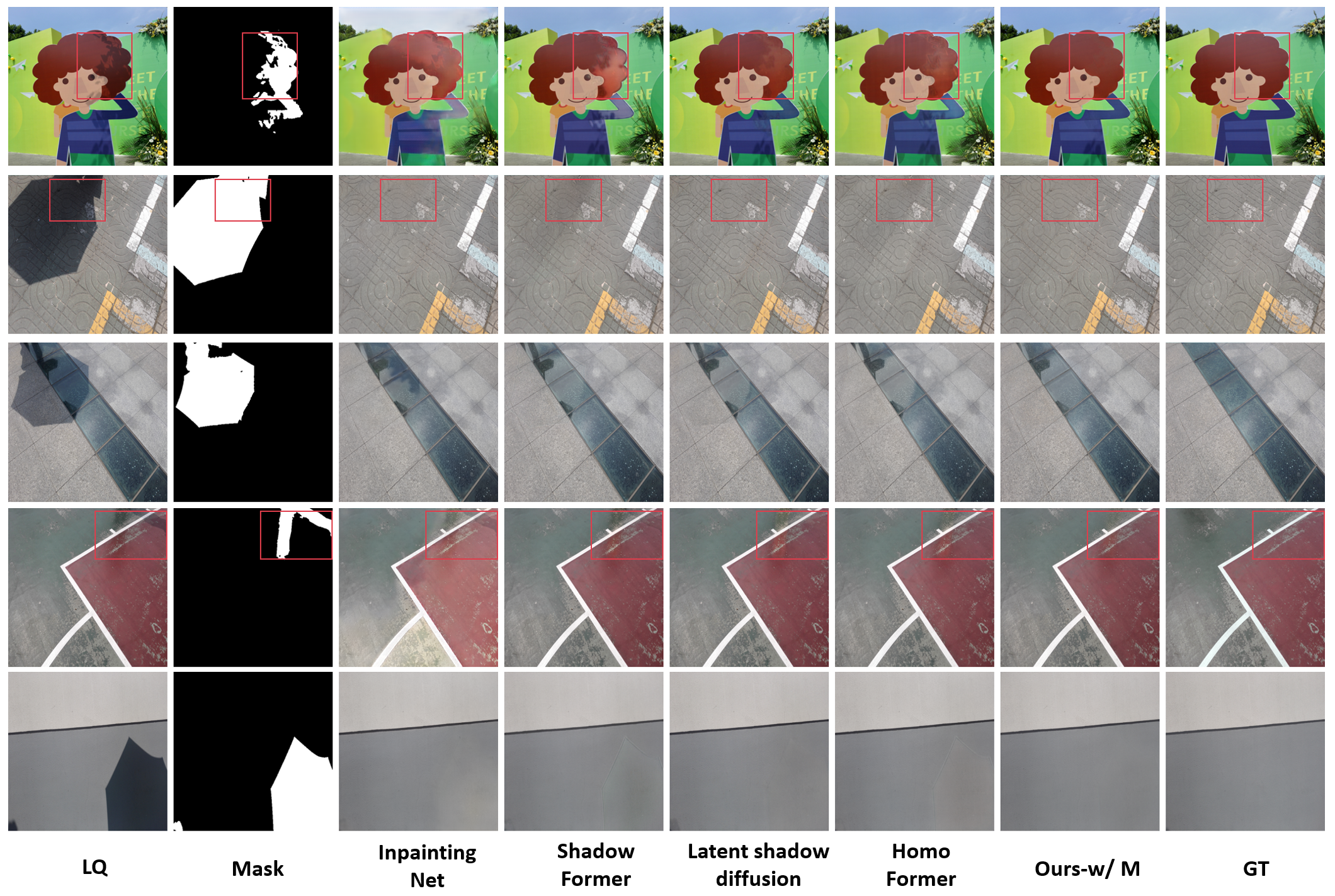}
    \caption{Visual comparison of various methods with input masks on the ISTD+ dataset.}
    \label{fig:istd+_mask}
\end{figure*}

 To highlight the advantages of the proposed method, we further extend our approach to explore its shadow removal performance under mask-available conditions and compare it with the SOTA shadow removal methods that utilize mask inputs. Specifically, we resize the mask and concatenate it with the original input of the control net before feeding it. We increase the number of channels of the input layer and initialize the new parameters to zero. We designate this version as Our-w/ M and conduct quantitative and qualitative evaluations of its outcomes on the ISTD+ dataset against existing mask-based shadow removal methods, including EPF Net \cite{fu2021auto}, SP+M+I Net \cite{le2021physics}, G2R Net \cite{liu2021shadow}, SG Net \cite{wan2022style}, BM Net \cite{zhu2022bijective}, Inpainting Net \cite{li2023leveraging}, ShadowFormer \cite{guo2023shadowformer}, Latent Shadow Diffusion \cite{mei2024latent}, and HomoFormer \cite{xiao2024homoformer}. 
As shown in Fig. \ref{fig:istd+_mask}, the proposed method achieves superior shadow removal performance. Compared to existing approaches, the method proposed in this paper leverages the generative priors of large models to effectively maintain consistency within and around shadows, with virtually no boundary artifacts. Furthermore, quantitative evaluation results listed in table \ref{table_ISTD+_mask} indicate that our method has comparable, or even superior PSNR and SSIM performance to the SOTA methods, validating the effectiveness of our proposed fidelity strategy. The outstanding LPIPS and FID scores further confirm the subjective performance superiority of our method, leading us to conclude that our approach surpasses existing methods in shadow removal capabilities.

\subsection{Ablation study}
\label{section4_3}
\begin{table*}[!t]
\label{SRD}
\renewcommand\arraystretch{1.3}
\caption{Ablation study and analysis on the proposed method.  Bold text indicates the best score}
\centering
\resizebox{350pt}{!}{\begin{tabular}{cccccccc}
\hline
                 \textbf{Method} & \textbf{PSNR}                & \textbf{PSNR-NS}              & \textbf{PSNR-S}             & \textbf{SSIM}                & \textbf{SSIM-NS}              & \textbf{SSIM-S}             & \textbf{LPIPS↓}                \\
\hline
\textbf{Ours full pipline} & \textbf{33.38 }                       & \textbf{36.41}                        & \textbf{37.92}                       & \textbf{0.964}                        & \textbf{0.977}                        & \textbf{0.991}                        & \textbf{0.0287}                       \\
\textbf{Ours-DDIM}    & 32.81                        & 35.75                        & 37.51 & 0.960  &                       0.974                       & 0.990                       & 0.0319 \\
\textbf{Ours-RDDM}    & 32.30                        & 35.53                        & 36.64 & 0.960  &                       0.974                       & 0.990                       & 0.0349 \\
\textbf{Ours-w/o EMA}    & 32.44                        & 35.94                        & 36.60 & 0.963  &                       0.977                       & 0.990                       & 0.0310 \\
\textbf{Ours-w/o detail-perserving decoder}    & 26.74                        & 28.00                        & 34.32 & 0.697  &                       0.760                      & 0.955                       & 0.1006 \\
\textbf{Ours-w/ SD-backbone}    & 32.56                        & 35.54                        & 37.33 & 0.960  &                    0.974                       & 0.990                       & 0.0348 \\
\hline
\end{tabular} 
}
\label{Ablation study and analysis}
\end{table*}

To substantiate the design choices and to further facilitate an understanding of our methodology, we conducted ablation experiments and analyses based on the ISTD+ dataset. 

\textbf{Noise and Shadow Residual Schedule.} To validate the effectiveness of the proposed shadow residual schedule, we made minimal modifications to our model to apply different schedules for predicting shadow-free images and conducted a quantitative assessment of the results. The methods adopting the DDIM and RDDM schedule are termed as Ours-DDIM and Ours-RDDM, respectively, with the corresponding results presented in Table \ref{Ablation study and analysis}. It should be noted that our strategy does not alter the original DDIM schedule but rather augments it with the shadow residual schedule. Therefore, for the DDIM version, we removed the NRD module as well as the proposed sampling strategy and executed DDIM sampling to facilitate the regeneration process from noise to image. Since its original strategy involves redesigning the noise-residual schedule, which necessitates an extra model branch for estimating the image residual, we employ the NRD module to produce the residual and noise. Subsequently, the sampling process of RDDM is implemented for the inference process. The results listed in Table \ref{Ablation study and analysis}  demonstrate that our strategy yields the highest PSNR and LPIPS scores, thereby confirming the enhancement in fidelity and subjective performance of our method compared to both the DDIM and RDDM strategies.

\textbf{Training Strategy.} 
To further verify the impact of the proposed self-enhancement training approach, an ablation test is also implemented. The quantitative results are presented in Table \ref{Ablation study and analysis} (Ours w/o EMA). It is evident that the proposed training strategy effectively enhances shadow removal performance. Because that it can reduce the reliance of the network on the results of previous time steps during the inference process, thereby minimizing the accumulation of errors.

\begin{figure}
    \centering
    \includegraphics[width=0.98\linewidth]{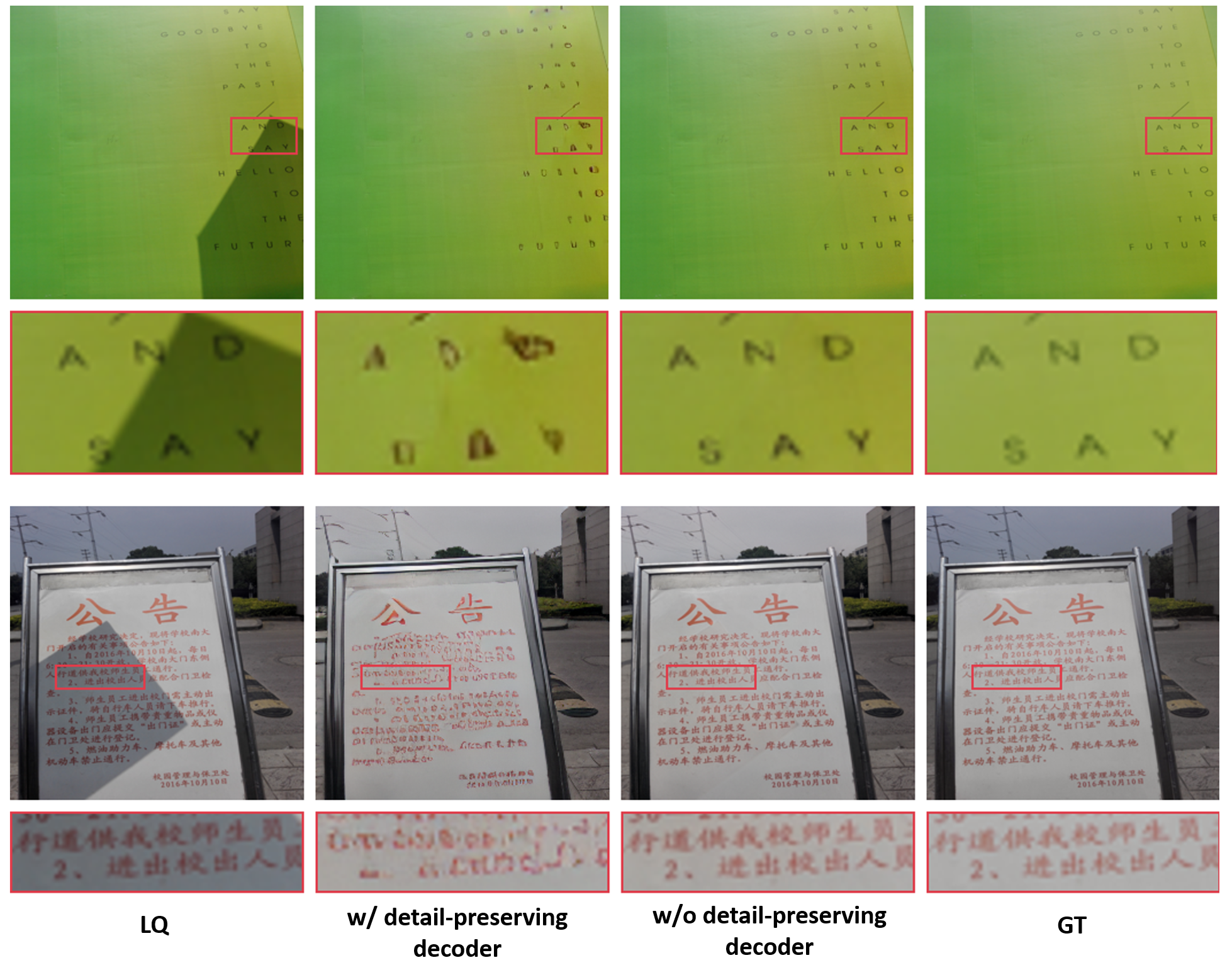}
    \caption{Comparison of visual results with and without the proposed detail preserved decoder in ablation study.}
    \label{fig:ablation-dec}
\end{figure}

\textbf{Decoder.} We present a comparative illustration in Fig. \ref{fig:ablation-dec}, showcasing the results obtained from the original decoder (w/o detail-preserving decoder) versus those from the fine-tuned decoder proposed in this paper, with a quantitative evaluation provided in Table \ref{Ablation study and analysis}. When utilizing the original decoder, there is a noticeable distortion of textual symbols in the image due to the loss of high-frequency details, making them challenging to discern. Conversely, the application of our fine-tuned decoder not only ensures the retention of high-frequency textures but also results in a marked improvement in the image's PSNR, SSIM, and LPIPS metrics. These improvements validate the effectiveness of our method in enhancing the fidelity of shadow removal results.

\textbf{Backbone Selection.} We explored the impact of various generative backbone models. Specifically, we replaced the backbone with a widely used generative network, the stable diffusion network (text-to-image generation model using v2-1-512-ema-pruned.ckpt), and presented the evaluation results in Table \ref{Ablation study and analysis} (ours w/ SD-backbone). Upon comparison, it can be observed that while the diffusion network based on the stable diffusion model also delivered excellent performance, the inpainting model adopted in our method outperformed it in terms of objective metrics. This superior performance is attributed to the inpainting model's approach of utilizing the latent representation of the shadow image and a full-zero mask as auxiliary inputs for the noisy image, which inherently preserves the details of the shadow image. Furthermore, the inpainting model also contributes to the enhancement of shadow removal as referenced in \cite{li2023leveraging}, thereby further improving the performance.

\section{Conclusion}
In this paper, a high-fidelity shadow removal scheme was proposed by means of a pre-trained large-scale generative model. To enhance fidelity during the generation process, we introduced a novel residual diffusion strategy on top of the conventional noise diffusion approach, aiming to generate the shadow residual component rather than a complete generation of the shadow-free image. Addressing the inconsistencies in input data between the training and inference phases of diffusion models, as well as the potential for error accumulation in the diffusion backward process, we presented a new training strategy that employs a model replica updated by an EMA strategy to augment the training data. Furthermore, a high-fidelity image encoder-decoder is designed. Extensive experiments demonstrate that the proposed method can both achieve higher visual performance than existing SOTA shadow removal approaches and obtain high fidelity by strictly preserving original contents in shadow regions.

 
%

\bibliography{reference}

\begin{thebibliography}{10}
\providecommand{\url}[1]{#1}
\csname url@samestyle\endcsname
\providecommand{\newblock}{\relax}
\providecommand{\bibinfo}[2]{#2}
\providecommand{\BIBentrySTDinterwordspacing}{\spaceskip=0pt\relax}
\providecommand{\BIBentryALTinterwordstretchfactor}{4}
\providecommand{\BIBentryALTinterwordspacing}{\spaceskip=\fontdimen2\font plus
\BIBentryALTinterwordstretchfactor\fontdimen3\font minus \fontdimen4\font\relax}
\providecommand{\BIBforeignlanguage}[2]{{%
\expandafter\ifx\csname l@#1\endcsname\relax
\typeout{** WARNING: IEEEtran.bst: No hyphenation pattern has been}%
\typeout{** loaded for the language `#1'. Using the pattern for}%
\typeout{** the default language instead.}%
\else
\language=\csname l@#1\endcsname
\fi
#2}}
\providecommand{\BIBdecl}{\relax}
\BIBdecl

\bibitem{guo2023shadowformer}
L.~Guo, S.~Huang, D.~Liu, H.~Cheng, and B.~Wen, ``Shadowformer: Global context helps shadow removal,'' in \emph{Proceedings of the AAAI Conference on Artificial Intelligence}, vol.~37, no.~1, 2023, pp. 710--718.

\bibitem{li2023leveraging}
X.~Li, Q.~Guo, R.~Abdelfattah, D.~Lin, W.~Feng, I.~Tsang, and S.~Wang, ``Leveraging inpainting for single-image shadow removal,'' in \emph{Proceedings of the IEEE/CVF International Conference on Computer Vision}, 2023, pp. 13\,055--13\,064.

\bibitem{liu2023shadow}
J.~Liu, Q.~Wang, H.~Fan, J.~Tian, and Y.~Tang, ``A shadow imaging bilinear model and three-branch residual network for shadow removal,'' \emph{IEEE Transactions on Neural Networks and Learning Systems}, pp. 1--15, 2023.

\bibitem{xiao2024homoformer}
J.~Xiao, X.~Fu, Y.~Zhu, D.~Li, J.~Huang, K.~Zhu, and Z.-J. Zha, ``Homoformer: Homogenized transformer for image shadow removal,'' in \emph{Proceedings of the IEEE/CVF Conference on Computer Vision and Pattern Recognition}, 2024, pp. 25\,617--25\,626.

\bibitem{le2019shadow}
H.~Le and D.~Samaras, ``Shadow removal via shadow image decomposition,'' in \emph{Proceedings of the IEEE/CVF International Conference on Computer Vision}, 2019, pp. 8578--8587.

\bibitem{ho2020denoising}
J.~Ho, A.~Jain, and P.~Abbeel, ``Denoising diffusion probabilistic models,'' \emph{Advances in neural information processing systems}, vol.~33, pp. 6840--6851, 2020.

\bibitem{song2020denoising}
J.~Song, C.~Meng, and S.~Ermon, ``Denoising diffusion implicit models,'' \emph{arXiv preprint arXiv:2010.02502}, 2020.

\bibitem{rombach2022high}
R.~Rombach, A.~Blattmann, D.~Lorenz, P.~Esser, and B.~Ommer, ``High-resolution image synthesis with latent diffusion models,'' in \emph{Proceedings of the IEEE/CVF conference on computer vision and pattern recognition}, 2022, pp. 10\,684--10\,695.

\bibitem{zhang2023adding}
L.~Zhang, A.~Rao, and M.~Agrawala, ``Adding conditional control to text-to-image diffusion models,'' in \emph{Proceedings of the IEEE/CVF International Conference on Computer Vision}, 2023, pp. 3836--3847.

\bibitem{TMMstyle}
Y.~Xu, X.~Xu, H.~Gao, and F.~Xiao, ``Sgdm: An adaptive style-guided diffusion model for personalized text to image generation,'' \emph{IEEE Transactions on Multimedia}, vol.~26, pp. 9804--9813, 2024.

\bibitem{TMManimegen}
Y.~Jiang, Q.~Liu, D.~Chen, L.~Yuan, and Y.~Fu, ``Animediff: Customized image generation of anime characters using diffusion model,'' \emph{IEEE Transactions on Multimedia}, pp. 1--13, 2024.

\bibitem{lin2023diffbir}
X.~Lin, J.~He, Z.~Chen, Z.~Lyu, B.~Dai, F.~Yu, W.~Ouyang, Y.~Qiao, and C.~Dong, ``Diffbir: Towards blind image restoration with generative diffusion prior,'' \emph{arXiv preprint arXiv:2308.15070}, 2023.

\bibitem{xie2023diffusion}
Y.~Xie, M.~Yuan, B.~Dong, and Q.~Li, ``Diffusion model for generative image denoising,'' \emph{arXiv preprint arXiv:2302.02398}, 2023.

\bibitem{kawar2022denoising}
B.~Kawar, M.~Elad, S.~Ermon, and J.~Song, ``Denoising diffusion restoration models,'' \emph{Advances in Neural Information Processing Systems}, vol.~35, pp. 23\,593--23\,606, 2022.

\bibitem{luo2023controlling}
Z.~Luo, F.~K. Gustafsson, Z.~Zhao, J.~Sj{\"o}lund, and T.~B. Sch{\"o}n, ``Controlling vision-language models for universal image restoration,'' \emph{arXiv preprint arXiv:2310.01018}, vol.~3, no.~8, 2023.

\bibitem{liu2024residual}
J.~Liu, Q.~Wang, H.~Fan, Y.~Wang, Y.~Tang, and L.~Qu, ``Residual denoising diffusion models,'' in \emph{Proceedings of the IEEE/CVF Conference on Computer Vision and Pattern Recognition}, 2024, pp. 2773--2783.

\bibitem{kuznetsova2020open}
A.~Kuznetsova, H.~Rom, N.~Alldrin, J.~Uijlings, I.~Krasin, J.~Pont-Tuset, S.~Kamali, S.~Popov, M.~Malloci, A.~Kolesnikov \emph{et~al.}, ``The open images dataset v4: Unified image classification, object detection, and visual relationship detection at scale,'' \emph{International journal of computer vision}, vol. 128, no.~7, pp. 1956--1981, 2020.

\bibitem{schuhmann2022laion}
C.~Schuhmann, R.~Beaumont, R.~Vencu, C.~Gordon, R.~Wightman, M.~Cherti, T.~Coombes, A.~Katta, C.~Mullis, M.~Wortsman \emph{et~al.}, ``Laion-5b: An open large-scale dataset for training next generation image-text models,'' \emph{Advances in Neural Information Processing Systems}, vol.~35, pp. 25\,278--25\,294, 2022.

\bibitem{RN5}
L.~Zhang, Q.~Zhang, and C.~Xiao, ``Shadow remover: Image shadow removal based on illumination recovering optimization,'' \emph{IEEE Transactions on Image Processing}, vol.~24, no.~11, pp. 4623--4636, 2015.

\bibitem{RN15}
Y.~Shor and D.~Lischinski, ``The shadow meets the mask: Pyramid‐based shadow removal,'' in \emph{Computer Graphics Forum}, vol.~27, 2008, pp. 577--586.

\bibitem{RN4}
G.~D. Finlayson, S.~D. Hordley, C.~Lu, and M.~S. Drew, ``On the removal of shadows from images,'' \emph{IEEE Transactions on Pattern Analysis and Machine Intelligence}, vol.~28, no.~1, pp. 59--68, 2005.

\bibitem{RN16}
M.~Gryka, M.~Terry, and G.~J. Brostow, ``Learning to remove soft shadows,'' \emph{ACM Transactions on Graphics}, vol.~34, no.~5, pp. 1--15, 2015.

\bibitem{RN6}
R.~Guo, Q.~Dai, and D.~Hoiem, ``Paired regions for shadow detection and removal,'' \emph{IEEE Transactions on Pattern Analysis and Machine Intelligence}, vol.~35, no.~12, pp. 2956--2967, 2012.

\bibitem{RN17}
T.~F.~Y. Vicente, M.~Hoai, and D.~Samaras, ``Leave-one-out kernel optimization for shadow detection and removal,'' \emph{IEEE Transactions on Pattern Analysis and Machine Intelligence}, vol.~40, no.~3, pp. 682--695, 2017.

\bibitem{wang2018stacked}
J.~Wang, X.~Li, and J.~Yang, ``Stacked conditional generative adversarial networks for jointly learning shadow detection and shadow removal,'' in \emph{Proceedings of the IEEE conference on computer vision and pattern recognition}, 2018, pp. 1788--1797.

\bibitem{fu2021auto}
L.~Fu, C.~Zhou, Q.~Guo, F.~Juefei-Xu, H.~Yu, W.~Feng, Y.~Liu, and S.~Wang, ``Auto-exposure fusion for single-image shadow removal,'' in \emph{Proceedings of the IEEE/CVF conference on computer vision and pattern recognition}, 2021, pp. 10\,571--10\,580.

\bibitem{TMMshaodwremoval}
K.~Niu, Y.~Liu, E.~Wu, and G.~Xing, ``A boundary-aware network for shadow removal,'' \emph{IEEE Transactions on Multimedia}, vol.~25, pp. 6782--6793, 2023.

\bibitem{jin2024des3}
Y.~Jin, W.~Ye, W.~Yang, Y.~Yuan, and R.~T. Tan, ``Des3: Adaptive attention-driven self and soft shadow removal using vit similarity,'' in \emph{Proceedings of the AAAI Conference on Artificial Intelligence}, vol.~38, no.~3, 2024, pp. 2634--2642.

\bibitem{hu2019mask}
X.~Hu, Y.~Jiang, C.-W. Fu, and P.-A. Heng, ``Mask-shadowgan: Learning to remove shadows from unpaired data,'' in \emph{Proceedings of the IEEE/CVF international conference on computer vision}, 2019, pp. 2472--2481.

\bibitem{cun2020towards}
X.~Cun, C.-M. Pun, and C.~Shi, ``Towards ghost-free shadow removal via dual hierarchical aggregation network and shadow matting gan,'' in \emph{Proceedings of the AAAI Conference on Artificial Intelligence}, vol.~34, no.~07, 2020, pp. 10\,680--10\,687.

\bibitem{jin2021dc}
Y.~Jin, A.~Sharma, and R.~T. Tan, ``Dc-shadownet: Single-image hard and soft shadow removal using unsupervised domain-classifier guided network,'' in \emph{Proceedings of the IEEE/CVF International Conference on Computer Vision}, 2021, pp. 5027--5036.

\bibitem{liu2023decoupled}
J.~Liu, Q.~Wang, H.~Fan, W.~Li, L.~Qu, and Y.~Tang, ``A decoupled multi-task network for shadow removal,'' \emph{IEEE Transactions on Multimedia}, vol.~25, pp. 9449--9463, 2023.

\bibitem{TMMinpainting}
C.~Zhang, W.~Yang, X.~Li, and H.~Han, ``Mmginpainting: Multi-modality guided image inpainting based on diffusion models,'' \emph{IEEE Transactions on Multimedia}, vol.~26, pp. 8811--8823, 2024.

\bibitem{TMMrestore}
Y.~Huang, J.~Huang, J.~Liu, M.~Yan, Y.~Dong, J.~Lv, C.~Chen, and S.~Chen, ``Wavedm: Wavelet-based diffusion models for image restoration,'' \emph{IEEE Transactions on Multimedia}, vol.~26, pp. 7058--7073, 2024.

\bibitem{guo2023shadowdiffusion}
L.~Guo, C.~Wang, W.~Yang, S.~Huang, Y.~Wang, H.~Pfister, and B.~Wen, ``Shadowdiffusion: When degradation prior meets diffusion model for shadow removal,'' in \emph{Proceedings of the IEEE/CVF Conference on Computer Vision and Pattern Recognition}, 2023, pp. 14\,049--14\,058.

\bibitem{mei2024latent}
K.~Mei, L.~Figueroa, Z.~Lin, Z.~Ding, S.~Cohen, and V.~M. Patel, ``Latent feature-guided diffusion models for shadow removal,'' in \emph{Proceedings of the IEEE/CVF Winter Conference on Applications of Computer Vision}, 2024, pp. 4313--4322.

\bibitem{yu2024scaling}
F.~Yu, J.~Gu, Z.~Li, J.~Hu, X.~Kong, X.~Wang, J.~He, Y.~Qiao, and C.~Dong, ``Scaling up to excellence: Practicing model scaling for photo-realistic image restoration in the wild,'' in \emph{Proceedings of the IEEE/CVF Conference on Computer Vision and Pattern Recognition}, 2024, pp. 25\,669--25\,680.

\bibitem{wu2024seesr}
R.~Wu, T.~Yang, L.~Sun, Z.~Zhang, S.~Li, and L.~Zhang, ``Seesr: Towards semantics-aware real-world image super-resolution,'' in \emph{Proceedings of the IEEE/CVF conference on computer vision and pattern recognition}, 2024, pp. 25\,456--25\,467.

\bibitem{yang2023paint}
B.~Yang, S.~Gu, B.~Zhang, T.~Zhang, X.~Chen, X.~Sun, D.~Chen, and F.~Wen, ``Paint by example: Exemplar-based image editing with diffusion models,'' in \emph{Proceedings of the IEEE/CVF Conference on Computer Vision and Pattern Recognition}, 2023, pp. 18\,381--18\,391.

\bibitem{he2020momentum}
K.~He, H.~Fan, Y.~Wu, S.~Xie, and R.~Girshick, ``Momentum contrast for unsupervised visual representation learning,'' in \emph{Proceedings of the IEEE/CVF conference on computer vision and pattern recognition}, 2020, pp. 9729--9738.

\bibitem{esser2021taming}
P.~Esser, R.~Rombach, and B.~Ommer, ``Taming transformers for high-resolution image synthesis,'' in \emph{Proceedings of the IEEE/CVF conference on computer vision and pattern recognition}, 2021, pp. 12\,873--12\,883.

\bibitem{zhu2023designing}
Z.~Zhu, X.~Feng, D.~Chen, J.~Bao, L.~Wang, Y.~Chen, L.~Yuan, and G.~Hua, ``Designing a better asymmetric vqgan for stablediffusion,'' \emph{arXiv preprint arXiv:2306.04632}, 2023.

\bibitem{zhu2019deformable}
X.~Zhu, H.~Hu, S.~Lin, and J.~Dai, ``Deformable convnets v2: More deformable, better results,'' in \emph{Proceedings of the IEEE/CVF conference on computer vision and pattern recognition}, 2019, pp. 9308--9316.

\bibitem{radford2021learning}
A.~Radford, J.~W. Kim, C.~Hallacy, A.~Ramesh, G.~Goh, S.~Agarwal, G.~Sastry, A.~Askell, P.~Mishkin, J.~Clark \emph{et~al.}, ``Learning transferable visual models from natural language supervision,'' in \emph{International conference on machine learning}.\hskip 1em plus 0.5em minus 0.4em\relax PMLR, 2021, pp. 8748--8763.

\bibitem{le2021physics}
H.~Le and D.~Samaras, ``Physics-based shadow image decomposition for shadow removal,'' \emph{IEEE Transactions on Pattern Analysis and Machine Intelligence}, vol.~44, no.~12, pp. 9088--9101, 2021.

\bibitem{RN7}
L.~Qu, J.~Tian, S.~He, Y.~Tang, and R.~W. Lau, ``Deshadownet: A multi-context embedding deep network for shadow removal,'' in \emph{Proceedings of the IEEE/CVF Conference on Computer Vision and Pattern Recognition}, 2017, pp. 4067--4075.

\bibitem{RN33}
R.~Zhang, P.~Isola, A.~A. Efros, E.~Shechtman, and O.~Wang, ``The unreasonable effectiveness of deep features as a perceptual metric,'' in \emph{Proceedings of the IEEE/CVF Conference on Computer Vision and Pattern Recognition}, 2018, pp. 586--595.

\bibitem{heusel2017gans}
M.~Heusel, H.~Ramsauer, T.~Unterthiner, B.~Nessler, and S.~Hochreiter, ``Gans trained by a two time-scale update rule converge to a local nash equilibrium,'' \emph{Advances in neural information processing systems}, vol.~30, 2017.

\bibitem{hu2018direction}
X.~Hu, L.~Zhu, C.-W. Fu, J.~Qin, and P.-A. Heng, ``Direction-aware spatial context features for shadow detection,'' in \emph{Proceedings of the IEEE conference on computer vision and pattern recognition}, 2018, pp. 7454--7462.

\bibitem{liu2021shadow}
Z.~Liu, H.~Yin, Y.~Mi, M.~Pu, and S.~Wang, ``Shadow removal by a lightness-guided network with training on unpaired data,'' \emph{IEEE Transactions on Image Processing}, vol.~30, pp. 1853--1865, 2021.

\bibitem{hu2024unveiling}
X.~Hu, Z.~Xing, T.~Wang, C.-W. Fu, and P.-A. Heng, ``Unveiling deep shadows: A survey on image and video shadow detection, removal, and generation in the era of deep learning,'' \emph{arXiv preprint arXiv:2409.02108}, 2024.

\bibitem{wan2022style}
J.~Wan, H.~Yin, Z.~Wu, X.~Wu, Y.~Liu, and S.~Wang, ``Style-guided shadow removal,'' in \emph{European Conference on Computer Vision}.\hskip 1em plus 0.5em minus 0.4em\relax Springer, 2022, pp. 361--378.

\bibitem{zhu2022bijective}
Y.~Zhu, J.~Huang, X.~Fu, F.~Zhao, Q.~Sun, and Z.-J. Zha, ``Bijective mapping network for shadow removal,'' in \emph{Proceedings of the IEEE/CVF Conference on Computer Vision and Pattern Recognition}, 2022, pp. 5627--5636.

\end{thebibliography}
\bibliographystyle{IEEEtran}

\end{document}